\documentclass{article}

\usepackage[numbers, compress]{natbib}
\usepackage{arxiv}

\PassOptionsToPackage{disable}{todonotes}

\usepackage[utf8]{inputenc}
\usepackage[T1]{fontenc}
\usepackage{hyperref}
\usepackage{url}
\usepackage{booktabs}
\usepackage{amsfonts}
\usepackage{amsmath}
\usepackage{nicefrac}
\usepackage{microtype}
\usepackage{xcolor}
\usepackage{graphicx}
\usepackage{enumitem}
\usepackage{caption}
\usepackage{subcaption}
\usepackage{float}
\usepackage{multirow}
\usepackage{colortbl}   
\usepackage{makecell}   
\usepackage{rotating}   
\usepackage{tikz}       
\usetikzlibrary{arrows.meta, positioning, shapes.geometric, calc}
\usepackage[most]{tcolorbox} 
\usepackage{fvextra}         
\usepackage{pifont}          
\usepackage[colorinlistoftodos,textsize=footnotesize,prependcaption]{todonotes}

\graphicspath{{figures/}}

\newif\ifneurips
\neuripsfalse

\newcommand{\dropyes}{\textcolor{green!55!black}{\ding{51}}}
\newcommand{\dropno}{\textcolor{red!70!black}{\ding{55}}}

\title{Reward-Free Evolving Agents via Pairwise Validator}

\author{%
  Minghao Liu\thanks{Corresponding author: \texttt{mh.liu.cs@gmail.com}} \quad
  Yu Wang \quad Jiayun Wang \quad Wei Wei \\
  Accenture
}

\newenvironment{ack}{\section*{Acknowledgments}}{}

\date{}

\begin{document}

\maketitle

\begin{abstract}
A \emph{self-evolving agentic loop} repeatedly proposes a tweaked version of an agent (its prompt template or program) and accepts or rejects the change based on a per-iteration quality signal. Designing that signal is often the costly part of the project: a reliable scalar reward requires domain expertise and labeled examples that are themselves as expensive to assemble as the agent's underlying task. We propose replacing the scalar at the accept/reject gate with a \emph{pairwise validator}: a frozen LLM that, given the parent and child candidate, returns a binary verdict on which is better. Pairwise judgment is generally easier and more stable than absolute scoring, due to its contrastive nature, which mitigates the need for strict scale calibration. The validator also requires no training of its own. We integrate the validator into three published self-evolving engines (GEPA, ADRS, ShinkaEvolve) and report two flavors: \emph{Adaptive Focus}, which retains the engine's existing val-set parent selection, and \emph{Soft Elo}, which lets the validator's verdicts drive parent selection so that val-set rewards drop as well. Across multiple agents and two artifact substrates (prompt and code), our method matches or exceeds the full-reward baseline on the majority of settings we evaluate, and the pattern survives a cross-family validator swap. The pairwise gate is thus a drop-in replacement for per-step reward design at competitive task accuracy without the labeling cost.
\end{abstract}

\section{Introduction}
\label{sec:intro}

Modern AI agents --- LLMs that solve multi-step tasks by combining a prompt, tool calls, and control logic --- are increasingly built and refined automatically rather than hand-tuned. One growing class of systems improves an agent \emph{without retraining the underlying model}: each iteration, the agent self-evolves by modifying the prompt template, the program, or the agentic workflow graph: a single piece of the agent's surrounding \emph{harness} or \emph{scaffolding}. The agent then runs this new version against a quality signal, and keeps the modification if the signal improves. Because the underlying model parameters remain frozen, such a self-evolution can be applied whether the model is open-weight or accessed only through an API. We refer to this class of systems as \emph{self-evolving agentic loops}. Existing literature along this line include GEPA \citep{gepa} for prompt evolution, ADRS \citep{adrs} and ShinkaEvolve \citep{shinkaevolve} for code evolution, and AlphaEvolve \citep{alphaevolve} and AFlow \citep{aflow} for related substrates. What unifies them is that the evolution progresses by referencing to only a scalar quality signal, namely a reward, which drives the per-iteration evolution direction and mutation selection. 

Building that scalar signal or reward for evolution is often considered expensive. Reward design takes domain expertise and frequently requires clean, labeled examples that are themselves costly to assemble --- for many real tasks, labeling a single example can cost more than running the agent itself. In some scenarios, such reward is hard to define or impossible to collect. The result is a practical asymmetry: collecting \emph{a few} examples of what good and bad outputs look like is often easy, but collecting \emph{enough} clean examples to define a reliable per-step reward is not. Meanwhile, recent work on RLAIF \citep{bai2022constitutional} and LLM-as-judge \citep{zheng2023llmasjudge} has shown that a frozen LLM, prompted with a handful of pairwise examples, can substitute for human preference labels in policy training. We ask the natural follow-up question: can a few-shot LLM validator substitute for the per-step task reward inside a self-evolving agentic loop's accept/reject decision?

\begin{figure*}[t]
  \centering
  \includegraphics[width=\linewidth]{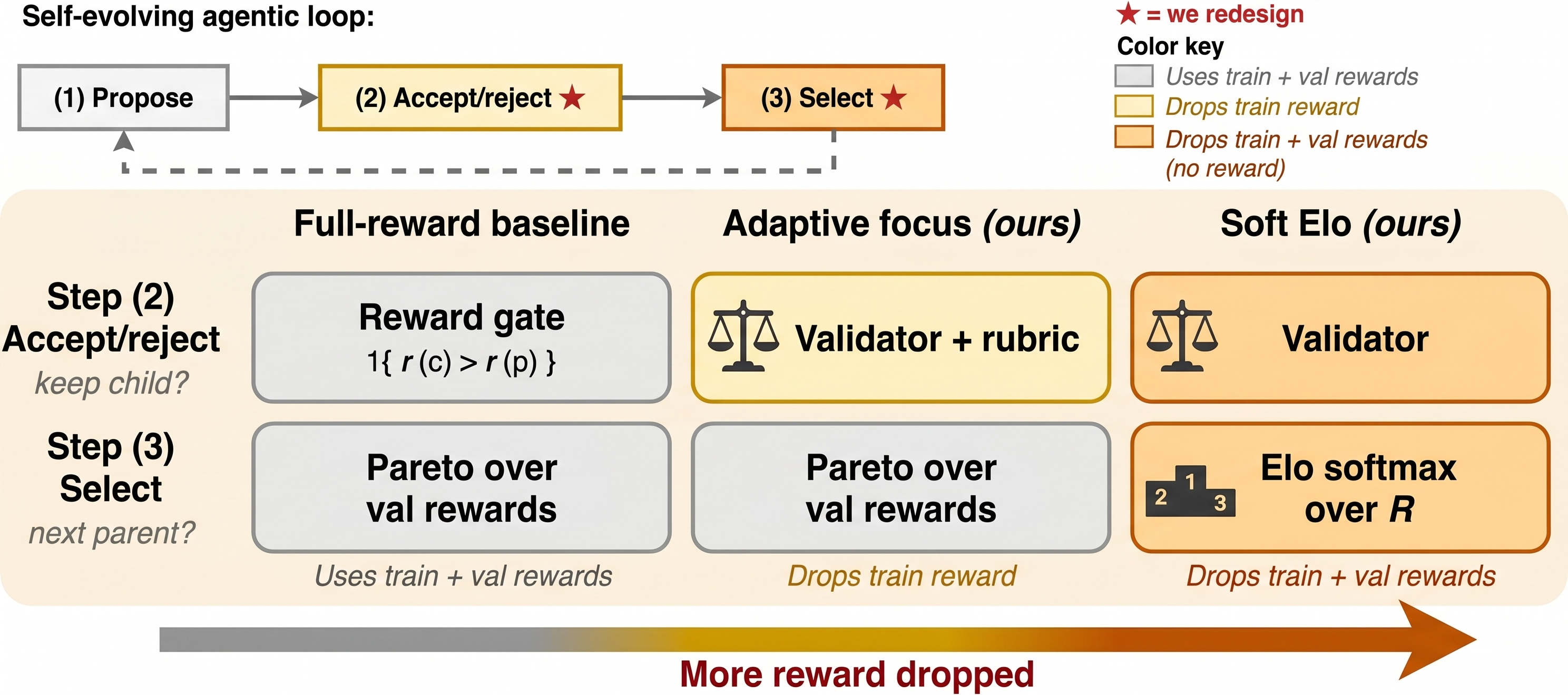}
  \vspace{-1.0em}
  \caption{\textbf{Overview of our framework.} A self-evolving agentic loop iterates over propose, accept-or-reject (the gate), and select. The gate traditionally uses a per-step task reward that compares parent and child prompt candidate scores. We replace that reward with a pairwise validator (LLM) that returns a binary verdict over parent and child. From left to right, the full-reward baseline (left) keeps both train and val rewards; \emph{Adaptive Focus} (center) swaps only the gate, dropping the train reward; \emph{Soft Elo} (right) further swaps parent selection, becoming reward-free.}
  \vspace{-1em}
  \label{fig:main}
\end{figure*}

Our answer is a \emph{pairwise validator} at the per-iteration accept/reject decision. Instead of scoring each candidate independently against a predefined task reward, we define an accept/reject gate that compares parent and child performance directly using a frozen LLM with a small number of in-context task examples (Figure~\ref{fig:main}). Pairwise comparison  judgment is generally easier and more stable than absolute scoring \citep{rlhf} --- the validator is asked the simpler \emph{which is better} question rather than to map outputs to a calibrated number --- and it requires no training of LLM model parameters. In this paper, we define two pairwise validator variants: {\textbf{\emph{Adaptive Focus}}} retains the engine's existing val-set parent selection while replacing only the gate; {\textbf{\emph{Soft Elo}}} additionally drives parent selection from the validator's decisions, so val-set rewards are also dropped. Across multiple agents and multiple engines spanning two artifact substrates (prompt evolution and code evolution), our method matches or exceeds the full-reward baseline on the majority of settings we evaluate. Our contributions are summarized below:
\begin{itemize}[leftmargin=1.2em,itemsep=0.1em,topsep=0.2em]
\item \textbf{Two variants of reward-free validation}. We replace the per-step task reward at the accept/reject gate with a validator (LLM) that returns a binary verdict over parent and child. \emph{Adaptive Focus} swaps only the gate and keeps the engine's val-reward parent selection. \emph{Soft Elo} further swaps parent selection for an Elo rating system driven by the validator's verdicts, dropping val-set rewards entirely. Both variants are training-free.
\item \textbf{Empirical superiority.} Our reward-free method matches or exceeds the full-reward state-of-the-art self-evolving agent baseline on most of the benchmarks we evaluate, across multiple agents and multiple engine setups under both prompt and code substrates. We also performed extensive ablation studies to investigate the mechanism behind such performance gain.
\end{itemize}

\section{Related Work}
\label{sec:related}

{\bf Pairwise feedback and LLM-as-judge}. Pairwise feedback can substitute for scalar rewards in two settings. In policy training, pairwise feedback replaces explicit scalar rewards: RLHF \citep{rlhf} learns a reward model from human pairwise preferences, RLAIF and Constitutional AI \citep{bai2022constitutional} substitute LLM-generated preferences for human ones with comparable downstream quality, and LLM-as-judge methods \citep{zheng2023llmasjudge} together with Chatbot Arena \citep{chatbotarena2024} show that frozen LLMs achieve high agreement with humans on pairwise comparisons across diverse tasks --- the property that makes them usable at the gate without an additional preference-learning step. In evolutionary search, non-scalar selection avoids reducing to a single fitness number: Quality-Diversity through AI / Human Feedback \citep{qdaif,lee2024qdhf} uses LLM- or human-derived similarity to maintain diversity niches, preference-based evolutionary multi-objective optimization \citep{huang2024prefemo} uses dueling bandits in lieu of scalar fitness, and multi-objective optimization itself \citep{nsgaii,moead} maintains a Pareto front over competing objectives. Both lines apply the preference signal over a pre-collected population. We instead run the validator online at the gate, on just the parent and child.

{\bf Self-evolving agent frameworks}. Self-evolving agent frameworks share a common loop: propose a candidate successor of an artifact (prompt, program, workflow), accept or reject the candidate against a quality signal, then select the next parent. GEPA \citep{gepa} optimizes prompts through reflective LLM-driven mutation with parent selection on a few-shot validation set. ADRS \citep{adrs} and ShinkaEvolve \citep{shinkaevolve} optimize code, the former through a parallel program database, the latter through reflective sampling with novelty pressure. AlphaEvolve \citep{alphaevolve} and OpenEvolve \citep{openevolve} apply analogous loops at larger scale. AFlow \citep{aflow} evolves agentic workflow graphs. The Darwin G\"odel Machine \citep{dgm} self-improves an entire agent codebase. Recent surveys consolidate the landscape \citep{gao2026survey,ecoffet2020openended}. Every published variant requires a scalar reward to drive selection. We replace that scalar reward with a pairwise validator at the accept/reject step.

{\bf Pairwise judges inside self-evolving agents}. The closest precedents combine pairwise judgment with self-evolving agents but stop short of substituting the per-step reward. CodeFavor \citep{liu2025codefavor} learns code preference models from synthetic evolution data but uses them outside the evolution loop. ExPairT-LLM \citep{expairt2026} pairwise-judges code candidates in a static benchmark setting. ScoreFlow \citep{wang2025scoreflow} uses pairwise preference inside agentic workflow optimization but on task performance, not as a reward substitute. MADE \citep{made2025} explores reward-free evolution through a different mechanism. Feedback Descent \citep{feedbackdescent2025} applies LLM textual feedback as a gradient-like signal outside the evolution loop, not at the gate. No prior system places a pairwise validator at the per-iteration accept/reject step as a replacement for the per-step task reward. That is the gap we address.

\section{Preliminaries}
\label{sec:prelims}

{\textbf{Self-evolving agentic loops}}. Each self-evolving agent refines a target component, such as a prompt or a piece of code, by iteratively proposing, accepting, and selecting candidates. The iteration generally follows three steps: \textbf{(1)~Propose.} Given a current ``parent'' candidate, the mutator generates a new ``child'' candidate by applying a modification. \textbf{(2)~Accept or reject.} A gate decides whether to keep the child by comparing parent and child against the engine's per-step task reward. \textbf{(3)~Select.} A selector picks the next parent from the pool for the next iteration. This paper redesigns the gate (step~{\textbf{(2)}}) and, in our \emph{Soft Elo} configuration, the selector (step~{\textbf{(3)}}); the propose step follows each engine's default. Figure~\ref{fig:main} contrasts the full-reward baseline (left) with our two proposed configurations \emph{Adaptive Focus} (center) and \emph{Soft Elo} (right) along the two redesigned steps.

\paragraph{Notation.}
Let \(f \in \mathcal{F}\) denote a candidate target component (a prompt
or a piece of code). At iteration \(t\), let \(f_{t,p} \in \mathcal{F}\)
denote the parent candidate selected by the engine from the candidate
pool, and let \(f_{t,c} \in \mathcal{F}\) denote the child candidate
proposed from this parent. The mutator defines a conditional
distribution over children given the parent:
\begin{equation}
  f_{t,c} \sim \mathcal{M}(\cdot \mid f_{t,p}),
  \label{eq:mutator}
\end{equation}
where \(\mathcal{M}\) is engine-specific 
(e.g., reflective LLM-driven editing in GEPA) and is treated as fixed
throughout this paper. Each engine also carries a per-step task reward
\(r : \mathcal{F} \to \mathbb{R}\), which maps a candidate to a scalar
quality value.

\paragraph{Data pools.}
We partition each benchmark's data into three disjoint pools. The validation pool $\mathcal{V}$, with $|\mathcal{V}| \in [12, 30]$ examples per benchmark, is deliberately smaller than the validation sets used in typical self-evolving setups --- we choose this low-data regime because labeling more examples is expensive. The full-reward baseline and our method use the same $\mathcal{V}$. The train pool $\mathcal{T}$ feeds the engine's mutator. In our method, the validator gate also draws a per-iteration minibatch $\mathcal{T}_t \subseteq \mathcal{T}$ from it. The remaining examples form a held-out test set used only for final reporting in \S\ref{sec:results}.

\paragraph{Full-reward baseline.}
The full-reward baseline runs each engine's default self-evolution loop. The gate (step~{\textbf{(2)}}) accepts the proposed child if its task reward exceeds the parent's:
\begin{equation}
  a_t^{\mathrm{full}}
  =
  \mathbf{1}\{\,r(f_{t,c}) > r(f_{t,p})\,\},
  \label{eq:baseline-gate}
\end{equation}
where \(\mathbf{1}\{\cdot\}\) is the indicator function. Accepted candidates are added to the engine's candidate pool. The selector (step~{\textbf{(3)}}) picks the next parent candidate \(f_{t+1,p}\) from this pool according to each engine's default rule (e.g., Pareto-style \citep{gepa,nsgaii}) over per-example validation rewards on \(\mathcal{V}\).

\section{Our Method}
\label{sec:method}
\label{sec:framework}

We replace the per-step task reward at the gate with a pairwise validator (\S\ref{sec:method-validator}). The two configurations differ in how aggressively they remove rewards: \emph{Adaptive Focus} (\S\ref{sec:method-adaptive}) drops the train reward at the gate but keeps validation rewards for parent selection; \emph{Soft Elo} (\S\ref{sec:method-elo}) drops both rewards and uses the validator's verdicts at both the gate and parent selection. Table~\ref{tab:conditions} lists all variants spanned by the two axes (\S\ref{sec:method-validator} for choices of $\mathcal{E}$; \S\ref{sec:method-elo} for the parent-selection rule); Appendix~\ref{app:elo-k-sweep} sweeps Soft Elo's step size $k$ further. Engine-specific integration details are deferred to Appendix~\ref{app:engine-integration}.

\begin{table}[t]
  \vspace{-1em}
  \centering
  \caption{\textbf{Method flavors and ablation variants.} The two flavors of our method (\textbf{bold}): \emph{Adaptive Focus} (\S\ref{sec:method-adaptive}) and \emph{Soft Elo} (\S\ref{sec:method-elo}). The remaining rows ablate two axes: \emph{what the validator sees} ($\mathcal{E}$) and \emph{how the Elo update is weighted}. The two right columns mark which reward inputs each \emph{family} drops relative to the full-reward baseline: \dropyes{} = drops it; \dropno{} = still uses it. }
  \label{tab:conditions}
  \small
  \setlength{\tabcolsep}{4pt}
  \begin{tabular}{@{}lp{0.46\linewidth}cc@{}}
    \toprule
    Condition & What the validator sees ($\mathcal{E}$) & \makecell{Drops train\\reward} & \makecell{Drops val\\rewards} \\
    \midrule
    Full-reward baseline                         & (no validator)                                                  & \dropno    & \dropno    \\
    \midrule
    \multicolumn{2}{l}{\textit{Train-reward-free family --- modifies step (2)}}                                                                                 & \dropyes   & \dropno    \\
    \quad Direct \textit{(\S\ref{sec:method-validator})}                                 & parent and child outputs only                                   &            &            \\
    \quad Few-shot (no rewards) \textit{(\S\ref{sec:method-validator})}                  & $+$ exemplars without reward labels                             &            &            \\
    \quad Few-shot (with rewards) \textit{(\S\ref{sec:method-validator})}                & $+$ exemplars with reward labels                                &            &            \\
    \quad \textbf{Adaptive Focus} \textit{(ours, \S\ref{sec:method-adaptive})} & self-adjusts rubric focus every 10 iters                       &            &            \\
    \midrule
    \multicolumn{2}{l}{\textit{Fully reward-free family --- modifies step (2) and step (3)}}                                                                              & \dropyes   & \dropyes   \\
    \quad Plain Elo \textit{(uniform, \S\ref{sec:method-elo})}          & uniform-step Elo updates ($c(V) \equiv 1$); used when the agent does not expose per-token logprobs\hspace{0pt} &            &            \\
    \quad \textbf{Soft Elo} \textit{(ours, \S\ref{sec:method-elo})}      & confidence-weighted Elo updates from validator verdicts         &            &            \\
    \bottomrule
  \end{tabular}
  \vspace{-1em}
\end{table}

\subsection{The Validator Gate}
\label{sec:method-validator}
\label{sec:framework-validator}

We propose a \emph{training-free} pairwise validator $V_\phi$ that replaces the explicit reward $r$ in the baseline gate rule (Eq.~\ref{eq:baseline-gate}). At the gate (step~(2)), the validator $V_\phi$ is queried with the parent and child along with a configuration-specific context $\mathcal{E}$:
\begin{equation}
  a_t^{V_\phi}
  =
  \mathbf{1}\{V_\phi(f_{t,c}, f_{t,p}, \mathcal{E}) = \texttt{better}\},
  \label{eq:validator-gate}
\end{equation}
where $V_\phi$ is a fixed LLM that returns a binary decision whether a child candidate is \texttt{better} or \texttt{worse} than its parent under greedy decoding ($T = 0$). The gate accepts on \texttt{better} (adding the child to the candidate pool) and rejects on \texttt{worse}. Per iteration, the validator returns one verdict per example in $\mathcal{T}_t$ ($|\mathcal{T}_t| = 3$ in our runs), and the gate accepts when the child wins at least as many verdicts as the parent.

For each train example, the validator sees the input plus parent and child outputs; the additional context $\mathcal{E}$ is the main design lever. Table~\ref{tab:conditions} lists the four choices we evaluate: \emph{Direct} ($\mathcal{E} = \emptyset$), two \emph{Few-shot} variants that append $N=3$ exemplars from $\mathcal{V}$ with or without reward labels, and \emph{Adaptive Focus} (\S\ref{sec:method-adaptive}), a self-adjusting rubric updated every 10 iterations. \emph{Soft Elo} (\S\ref{sec:method-elo}) keeps $\mathcal{E} = \emptyset$ but additionally swaps the parent-selection rule.

\subsection{Adaptive Focus (training-reward-free)}
\label{sec:method-adaptive}

Adaptive Focus extends the validator gate by making the context $\mathcal{E}$ a periodically-refreshed rubric handed to $V_\phi$ at gate time; the rubric is the only thing this configuration changes. Parent selection (step~{\textbf{(3)}}) follows the engine's default rule over validation rewards on $\mathcal{V}$.

\paragraph{Three rubrics.} The rubric is one of three pre-defined options, each broader than the previous:
\begin{enumerate}\setlength{\itemsep}{0pt}
  \item \emph{Accuracy}: ask which response has the better final answer.
  \item \emph{Soundness}: ask which response gives the better answer with a clear reasoning chain.
  \item \emph{Holistic}: ask which response is stronger overall, weighing correctness, reasoning, completeness, clarity, and formatting together.
\end{enumerate}
At gate time, the validator selects one of these three rubrics as context $\mathcal{E}$, directing focus to a different dimension of the pairwise decision.

\paragraph{Periodic rubric refresh.} The active rubric is refreshed every $10$ iterations by a meta-LLM call. The meta-LLM is shown a compact window of recent gate activity --- gate decisions over the last $10$ iterations and a small set of recent (parent, child) output snippets --- and outputs one of the three rubrics for the next window. The rubric is initialized at \emph{Accuracy}; there is no fixed schedule, though we empirically observe that the rubric tends to shift toward \emph{Holistic} over the course of a run.

\subsection{Soft Elo (Fully Reward-Free)}
\label{sec:method-elo}

Soft Elo extends the validator gate from Eq. (\ref{eq:validator-gate}) with $\mathcal{E} = \emptyset$, and replaces the engine's default parent-selection rule (step~(3)) with an Elo rating system \citep{elo1978, chatbotarena2024} over pairwise verdicts. Unlike the default rule, which selects on validation rewards over $\mathcal{V}$, the Elo updates here are driven by validator verdicts on the same train minibatch $\mathcal{T}_t$ that step (2) consumes; no validation rewards are used. Soft Elo is therefore completely reward-free during evolution.

Each candidate in the engine's candidate pool carries a rating $R$, initialized at $1600$. At iteration $t$, the engine samples a parent $f_{t,p}$ from a softmax over the pool's current ratings, and the mutator proposes a child $f_{t,c}$. The child is matched against the parent and against three randomly sampled members of the pool; each match runs the validator $V_\phi$ on every example $x$ in the train minibatch $\mathcal{T}_t$, returning a per-example verdict $S(x) \in \{0, \tfrac{1}{2}, 1\}$ (opponent better, tie, child better). Every verdict triggers a paired Elo update on the two contestants. Letting $R_c$ and $R_o$ denote the child's and the opponent's ratings, the child's expected score and rating update are
\begin{equation}
  \hat S_c \;=\; \frac{1}{1 + 10^{(R_o - R_c)/400}}, \qquad R_c \;\leftarrow\; R_c + k\,c_t\,\bigl(S(x) - \hat S_c\bigr),
  \label{eq:elo-update}
\end{equation}
and $R_o$ is updated symmetrically. Here $k = 32$ is the base step and $c_t \in [0, 1]$ is a per-iteration confidence weight: the mean per-token probability of the agent's previous-iteration response, $c_t = \tfrac{1}{|y|}\sum_{i=1}^{|y|} p(y_i \mid y_{<i}, x)$ for response tokens $y = (y_1, \ldots, y_{|y|})$. The same scalar is shared across all verdicts at iteration~$t$. After all $4\,|\mathcal{T}_t|$ updates the child is admitted to the pool at its current rating. Appendix~\ref{app:elo-k-sweep} sweeps $k$ further.

When the agent does not expose per-token probabilities --- so $c_t$ cannot be computed --- we fall back to \emph{Plain Elo} (Table~\ref{tab:conditions}), which keeps the same setup but with $c_t = 1$ and a larger step $k = 64$ to partly offset the missing confidence weighting.

\section{Results}
\label{sec:results}

Replacing the per-step reward with a pairwise validator does not cost us on task accuracy: across multiple agents and engines, our method matches or exceeds the full-reward baseline on the majority of cells we measure (\S\ref{sec:results-nlp}), and on cells where validation saturates early it shows a smaller validation-to-test gap than the baseline. The pattern survives a cross-family validator swap (\S\ref{sec:results-judge-swap}) and reproduces on cross-substrate cells under code evolution (\S\ref{sec:results-code}). Reaching the baseline \emph{is} the result: we removed the per-step reward and the loop's task score survived.

\subsection{Setup}
\label{sec:setup}
\label{sec:method-tasks}

\paragraph{Tasks.}
Ten tasks across three families: \textbf{retrieval / instruction-following} --- HotpotQA \citep{hotpotqa}, HoVer \citep{hover}, IFBench \citep{ifbench}, PUPA \citep{pupa}; \textbf{math} --- AIME \citep{aime} and LiveBench-Math \citep{livebench}; \textbf{code evolution} --- \texttt{circle\_packing}, \texttt{signal\_processing}, and \texttt{txn\_scheduling} \citep{adrs}, plus \texttt{julia\_prime\_counting} \citep{shinkaevolve}. The first six tasks are evolved with GEPA; the four code tasks with ADRS and ShinkaEvolve, with train / validation / test splits and budgets following each engine's published recipe.

\paragraph{Full-reward baselines.} For each substrate we compare against the engine's published full-reward configuration. \textbf{GEPA}~\citep{gepa} evolves prompts through reflective LLM-driven mutation with Pareto parent selection over per-example validation rewards. \textbf{ADRS}~\citep{adrs} evolves code through a parallel program database that selects parents via the full task evaluator. \textbf{ShinkaEvolve}~\citep{shinkaevolve} evolves code through reflective sampling with novelty pressure, again gated and selected on the engine's task reward. Our methods replace only the gate (and, for Soft Elo, the parent-selection rule); the mutator and the engine's default scoring stay fixed.

\textbf{Models.}
We report five agent configurations: Qwen3-8B, Qwen3-8B (thinking), Qwen3-32B, Gemma-4, and gpt-oss-20b. Per-row caveats (saturation, math-floor regimes) are disclosed inline at each results subsection.

\textbf{Reporting protocol.}
For each \texttt{(model, task, method)} cell we report the test-set score at the best-validation checkpoint; per-iteration trajectories appear in Figure~\ref{fig:qa-trajectories}.

\subsection{Main Results on Prompt Evolution}
\label{sec:results-nlp}

\begin{figure}[t]
  \centering
  \vspace{-1em}
  \includegraphics[width=0.9\linewidth]{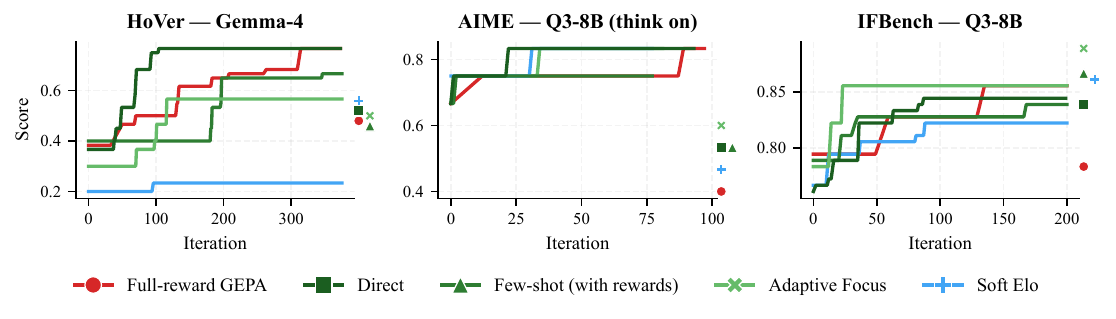}
  \vspace{-0.8 em}
  \caption{\textbf{Per-iteration GEPA trajectories where the validation-to-test gap is most visible.}
    Panels: HoVer (Gemma-4), AIME (Qwen3-8B (thinking)), IFBench (Qwen3-8B). Solid line = running validation-best trajectory; right-edge marker = per-variant best test score (matches cells in Tables~\ref{tab:combined-main-nlp} and~\ref{tab:combined-main-math}). Color: red full-reward GEPA; green train-reward-free family (Direct, Few-shot (with rewards), Adaptive Focus); blue fully reward-free (Soft Elo). See \S\ref{sec:results-nlp} for the gap discussion; \S\ref{sec:prelims} for our deliberate small validation pool ($|\mathcal{V}| \in [12, 30]$).}
  \vspace{-1.0 em}
  \label{fig:qa-trajectories}
\end{figure}

We evaluate five agent configurations on the four retrieval and instruction-following tasks (Table~\ref{tab:combined-main-nlp}) with GEPA as the full-reward baseline. Throughout this section the validator $V_\phi$ and the agent share the same LLM (self-paired); per-iteration trajectories are shown in Figure~\ref{fig:qa-trajectories}.

\begin{table}[t]
  \centering
  \caption{\textbf{Per-model GEPA results on retrieval / instruction-following tasks.} The per-row best across the validator slate matches or exceeds full-reward GEPA on all $20$ rows. Cell = test score at the best-validation checkpoint; bold = per-row best. Columns split into two families: train-reward-free (Direct, Few-shot $\times 2$, Adaptive Focus) and fully reward-free (Plain Elo, Soft Elo). \textit{(ours)} marks Adaptive Focus and Soft Elo; \textit{(uniform)} marks Plain Elo, the fallback when per-token probabilities are unavailable. Metric: Accuracy (\%).}
  \label{tab:combined-main-nlp}
  \small
  \setlength{\tabcolsep}{4pt}
  \resizebox{\columnwidth}{!}{%
  \begin{tabular}{@{}llccccccc@{}}
    \toprule
    \multirow{2}{*}{Model} & \multirow{2}{*}{Task}
      & \multirow{2}{*}{\makecell{Full-reward\\GEPA}}
      & \multicolumn{4}{c}{Train-reward-free}
      & \multicolumn{2}{c}{Fully reward-free} \\
    \cmidrule(lr){4-7}\cmidrule(lr){8-9}
      &       &       & Direct & \makecell{Few-shot\\(no rewards)} & \makecell{Few-shot\\(with rewards)} & \makecell{\textbf{Adaptive Focus}\\\textit{(ours)}} & \makecell{Plain Elo\\\textit{(uniform)}} & \makecell{\textbf{Soft Elo}\\\textit{(ours)}} \\
    \midrule
      \multirow{4}{*}{\shortstack[l]{Qwen3-8B}} & HotpotQA & 77.0 & 77.2 & 73.5 & 75.5 & \textbf{78.5} & 64.3 & 59.3 \\
       & IFBench & 78.3 & 83.9 & 78.9 & 86.7 & \textbf{88.9} & 85.6 & 86.1 \\
       & HoVer & 50.0 & 50.0 & 52.0 & 54.0 & 52.0 & 58.0 & \textbf{60.0} \\
       & PUPA & 62.1 & 60.6 & 60.9 & 57.0 & \textbf{63.6} & 58.7 & 58.7 \\
    \midrule
      \multirow{4}{*}{\shortstack[l]{Qwen3-8B (thinking)}} & HotpotQA & 77.4 & 78.3 & 80.6 & \textbf{82.1} & 79.2 & 81.0 & 75.0 \\
       & IFBench & 82.8 & 86.7 & 79.4 & \textbf{92.2} & 89.2 & 88.3 & 82.2 \\
       & HoVer & 52.0 & 51.0 & 51.5 & 52.0 & \textbf{64.0} & 52.0 & 58.0 \\
       & PUPA & 79.2 & 78.2 & 77.4 & \textbf{80.2} & 66.0 & 61.4 & 64.6 \\
    \midrule
      \multirow{4}{*}{\shortstack[l]{Qwen3-32B}} & HotpotQA & 84.4 & 88.7 & 87.3 & \textbf{91.3} & 88.0 & 83.1 & 87.8 \\
       & IFBench & 75.6 & 82.8 & 82.2 & 83.3 & \textbf{87.2} & 83.9 & 80.6 \\
       & HoVer & 58.0 & 52.0 & 52.0 & 56.0 & \textbf{60.0} & 56.0 & 58.0 \\
       & PUPA & 66.8 & 67.6 & 65.2 & 68.8 & 68.5 & \textbf{76.0} & 67.8 \\
    \midrule
      \multirow{4}{*}{\shortstack[l]{Gemma-4}} & HotpotQA & 75.7 & \textbf{79.0}& 71.9 & 75.5  & 78.9 & 78.4 & 73.7 \\
       & IFBench & 88.9 & 84.3 & 73.9 & 85.6 & 91.7 & 89.4 & \textbf{93.9} \\
       & HoVer & 48.0 & 52.0 & 46.0 & \textbf{58.0} & 56.2 & 56.0 & \textbf{58.0} \\
       & PUPA & 71.8 & 69.2 & 71.9 & 72.2 & 73.1 & \textbf{75.4} & 72.2 \\
    \midrule
      \multirow{4}{*}{\shortstack[l]{gpt-oss-20b}} & HotpotQA & 86.9 & 80.3 & 88.9 & 84.4 & 80.1 & 86.9 & \textbf{89.1} \\
       & IFBench & 86.7 & 87.2 & 83.9 & \textbf{91.7} & \textbf{91.7} & 81.7 & 87.2 \\
       & HoVer & 64.0 & 56.0 & 60.0 & \textbf{68.0} & 62.0 & 58.0 & 58.0 \\
       & PUPA & 58.8 & 62.7 & 63.2 & \textbf{72.3} & 68.8 & 65.1 & 66.4 \\
    \bottomrule
  \end{tabular}}%
  \vspace{-1 em}
\end{table}

In Table~\ref{tab:combined-main-nlp}, our two proposed methods (\emph{Adaptive Focus} and \emph{Soft Elo}) together with the three context-only variants (\emph{Direct}, \emph{Few-shot (no rewards)}, \emph{Few-shot (with rewards)}) match or outperform the full-reward GEPA baseline across the four retrieval and instruction-following tasks and five agent configurations. This demonstrates the viability of eliminating reward signals during training. Three illustrative wins span agent families, task types, and winning-variant identities: \emph{Soft Elo} on HoVer (Qwen3-8B, \textcolor{green!55!black}{$+10.0$}), \emph{Adaptive Focus} on IFBench (Qwen3-32B, \textcolor{green!55!black}{$+11.6$}), and \emph{Few-shot (with rewards)} on PUPA (gpt-oss-20b, \textcolor{green!55!black}{$+13.5$}). \emph{Few-shot (with rewards)} is the strongest single ablation, taking the per-row best on $8$ of $20$ cells; this shows that supplying labeled exemplars to the validator is a competitive alternative to the per-task rubric rotation when a small set of reward labels is available. Within our two flagship methods, \emph{Adaptive Focus} tends to win where the gate's task focus matters most (e.g., IFBench), while \emph{Soft Elo} edges out where the validation pool is noisy enough that Elo's pool-wide smoothing helps over default Pareto selection (e.g., HoVer).

We compare generalization across methods. In Figure~\ref{fig:qa-trajectories}, solid lines denote the running validation-best trajectory and right-edge markers are per-variant best test scores. Our pairwise validator variants --- \emph{Few-shot (with rewards)}, \emph{Adaptive Focus}, and \emph{Soft Elo} --- show a smaller validation-to-test gap than the full-reward GEPA on cells where validation saturates early.
On the three cells where the baseline saturates validation early but lags on test --- HoVer (Gemma-4), AIME (Qwen3-8B (thinking)), IFBench (Qwen3-8B) --- our methods recover the gap: \emph{Soft Elo} lifts HoVer by \textcolor{green!55!black}{$+10.0$}; on AIME, \emph{Adaptive Focus} and \emph{Plain Elo} tie the per-row best at \textcolor{green!55!black}{$+20.0$} (three problems); \emph{Adaptive Focus} lifts IFBench by \textcolor{green!55!black}{$+10.6$}. AIME and IFBench show this validation-to-test gap most visibly: full-reward GEPA's validation trajectory climbs to the top of the panel while its test marker stays well below. We read this as a property of our deliberate low-data regime ($|\mathcal{V}| \in [12, 30]$, \S\ref{sec:prelims}) rather than a flaw in GEPA itself --- with a small validation pool, validation rewards are a noisier proxy for test performance, and any engine that selects on $\mathcal{V}$ can be pulled toward whatever moves the pool. Our pairwise gate does not depend on $\mathcal{V}$ at decision time and is therefore less sensitive to this regime. We attribute the generalization advantage to the pairwise comparison formulation, which does not expose an explicit reward during training, and conjecture that pairwise comparison carries generalization signal that scalar validation or training rewards do not.

\textbf{Math results (robustness check).}
\label{sec:results-math}
On math (Table~\ref{tab:combined-main-math}, AIME and LiveBench-Math against GEPA), we treat results as a robustness check: AIME has $15$ test problems and LiveBench-Math has $30$, so a single solved problem corresponds to $6.7\%$ or $3.3\%$ of the score. The clearest signal is Qwen3-8B (thinking) on AIME, where \emph{Few-shot (with rewards)}, \emph{Adaptive Focus}, and \emph{Plain Elo} each lift accuracy from $40.0\%$ to $60.0\%$ (\textcolor{green!55!black}{$+20.0$}, three problems). Other cells sit at parity or within one solved problem of the baseline; on Gemma-4, \emph{Soft Elo} adds one problem on LB-Math (\textcolor{green!55!black}{$+3.3$}). The pattern is consistent with prompt evolution --- our methods at parity or above where the regime allows --- though math's small test-set size limits the strength of the claim.

\begin{table}[t]
  \centering
  \caption{\textbf{Per-model GEPA results on math tasks (robustness check).} Same column slate as Table~\ref{tab:combined-main-nlp}, on AIME (15 problems) and LiveBench-Math (30); deltas are single-problem-granularity ($6.7\%$ on AIME, $3.3\%$ on LB-Math) in a hard regime. Qwen3-8B, Qwen3-32B, and gpt-oss-20b excluded: scores floor at $0$ or sit too low. Bold = per-row best.}
  \label{tab:combined-main-math}
  \small
  \setlength{\tabcolsep}{4pt}
  \resizebox{\columnwidth}{!}{%
  \begin{tabular}{@{}llccccccc@{}}
    \toprule
    \multirow{2}{*}{Model} & \multirow{2}{*}{Task}
      & \multirow{2}{*}{\makecell{Full-reward\\GEPA}}
      & \multicolumn{4}{c}{Train-reward-free}
      & \multicolumn{2}{c}{Fully reward-free} \\
    \cmidrule(lr){4-7}\cmidrule(lr){8-9}
      &       &       & Direct & \makecell{Few-shot\\(no rewards)} & \makecell{Few-shot\\(with rewards)} & \makecell{\textbf{Adaptive Focus}\\\textit{(ours)}} & \makecell{Plain Elo\\\textit{(uniform)}} & \makecell{\textbf{Soft Elo}\\\textit{(ours)}} \\
    \midrule
      \multirow{2}{*}{\shortstack[l]{Qwen3-8B (thinking)}} & AIME & 40.0 & 53.3 & 53.3 & \textbf{60.0} & \textbf{60.0} & \textbf{60.0} & 46.7 \\
       & LB-Math & \textbf{33.3} & \textbf{33.3} & 26.7 & 26.7 & \textbf{33.3} & 26.7 & 23.3 \\
    \midrule
      \multirow{2}{*}{\shortstack[l]{Gemma-4}} & AIME & 26.7 & \textbf{33.3} & 26.7 & \textbf{33.3} & \textbf{33.3} & 26.7 & 26.7 \\
       & LB-Math & 33.3 & 30.0 & 30.0 & 33.3 & 30.0 & 33.3 & \textbf{36.7} \\
    \bottomrule
  \end{tabular}}%
  \vspace{-1em}
\end{table}

\subsection{Validator Ablation on Prompt Evolution}
\label{sec:results-judge-swap}

To confirm that our results are not a self-evaluation artifact --- where the same model both proposes the candidate and judges it --- we keep the agent model fixed at Qwen3-8B and swap the validator $V_\phi$ for Claude Haiku 4.5 or Gemma-4 (Table~\ref{tab:ablations-judge}). Since the agent is unchanged, the full-reward GEPA column is inherited from Table~\ref{tab:combined-main-nlp}; the remaining columns report our proposed methods under the swapped $V_\phi$.

\begin{table}[t]
  \centering
  \caption{\textbf{Validator-swap ablation.} Same agent as the headline (Qwen3-8B) but the validator is replaced with Claude Haiku 4.5 (top) and Gemma-4 (bottom). The full-reward GEPA column is inherited from Table~\ref{tab:combined-main-nlp} since the agent is unchanged. \textit{(ours)} marks Adaptive Focus and Soft Elo; \textit{(uniform)} marks Plain Elo. Bold cells = per-row winner across baseline plus all variants.}
  \label{tab:ablations-judge}
  \small
  \setlength{\tabcolsep}{4pt}
  \resizebox{\columnwidth}{!}{%
  \begin{tabular}{@{}llccccccc@{}}
    \toprule
    \multirow{2}{*}{Model} & \multirow{2}{*}{Task}
      & \multirow{2}{*}{\makecell{Full-reward\\GEPA}}
      & \multicolumn{4}{c}{Train-reward-free}
      & \multicolumn{2}{c}{Fully reward-free} \\
    \cmidrule(lr){4-7}\cmidrule(lr){8-9}
      &       &       & Direct & \makecell{Few-shot\\(no rewards)} & \makecell{Few-shot\\(with rewards)} & \makecell{\textbf{Adaptive Focus}\\\textit{(ours)}} & \makecell{Plain Elo\\\textit{(uniform)}} & \makecell{\textbf{Soft Elo}\\\textit{(ours)}} \\
    \midrule
      \multirow{4}{*}{\shortstack[l]{Qwen3-8B \\ (with Haiku4.5\\ validator)}} & HotpotQA & 77.0 & 70.6 & 76.8 & 77.7 & \textbf{80.5} & 71.6 & 59.1 \\
       & IFBench & 78.3 & 85.6 & 86.1 & \textbf{87.2} & 86.9 & 86.7 & 86.7 \\
       & HoVer & 50.0 & 54.0 & 56.0 & \textbf{57.1} & 55.0 & 52.0 & 52.0 \\
       & PUPA & 62.1 & 65.0 & \textbf{69.1} & 55.1 & 61.0 & 67.2 & 63.0 \\
    \midrule
      \multirow{4}{*}{\shortstack[l]{Qwen3-8B \\ (with Gemma\\ validator)}} & HotpotQA & 77.0 & \textbf{78.2} & 75.6 & 77.6 & 76.4 & 61.8 & 60.3 \\
       & IFBench & 78.3 & 84.4 & 85.0 & 85.6 & 83.9 & 82.8 & \textbf{88.3} \\
       & HoVer & 50.0 & 56.0 & \textbf{60.0} & 54.0 & 56.0 & 54.0 & 53.0 \\
       & PUPA & \textbf{62.1} & 60.1 & 53.5 & 59.3 & 59.3 & 60.1 & 55.0 \\
    \bottomrule
  \end{tabular}}%
\end{table}

Under the Haiku validator, the per-row best across our proposed variants matches or exceeds full-reward GEPA on all four tasks: \emph{Adaptive Focus} on HotpotQA (\textcolor{green!55!black}{$+3.5$}), \emph{Few-shot (with rewards)} on IFBench (\textcolor{green!55!black}{$+8.9$}) and HoVer (\textcolor{green!55!black}{$+7.1$}), and \emph{Few-shot (no rewards)} on PUPA (\textcolor{green!55!black}{$+7.0$}). Under Gemma-4, three of four tasks meet or beat the baseline: \emph{Direct} on HotpotQA (\textcolor{green!55!black}{$+1.2$}), \emph{Soft Elo} on IFBench (\textcolor{green!55!black}{$+10.0$}), and \emph{Few-shot (no rewards)} on HoVer (\textcolor{green!55!black}{$+10.0$}); PUPA falls short at \textcolor{red!70!black}{$-2.0$} (top variant $60.1$ vs.\ baseline $62.1$). The pattern broadly transfers across two validator families from outside the agent's model family, suggesting that our pairwise-validator gains are not a self-pairing artifact but reflect a genuine benefit from removing explicit reward at training and validation time. The Gemma-4 PUPA shortfall is the lone exception we surface in this swap; we revisit this case in the limitations.

\subsection{Main Results on Code Evolution}
\label{sec:results-code}

We test the pairwise validator on code evolution under ADRS \citep{adrs} and ShinkaEvolve \citep{shinkaevolve}, with Qwen3-8B as the agent model. ADRS covers three tasks (\texttt{txn\_scheduling}, \texttt{circle\_packing}, \texttt{signal\_processing}); ShinkaEvolve covers two (\texttt{circle\_packing}, \texttt{julia\_prime\_counting}). Each cell is evaluated under two validator setups: self-paired (Qwen3-8B as both agent and $V_\phi$) and cross-family (Qwen3-8B as agent, Claude Haiku 4.5 as $V_\phi$). Table~\ref{tab:code-deltas} reports the test-set scores.

\textbf{Why a different variant slate.}
The variant slate in Table~\ref{tab:code-deltas} differs from the prompt-evolution slate of \S\ref{sec:results-nlp} for an engine-side reason: ADRS and ShinkaEvolve do not maintain a held-out validation pool $\mathcal{V}$ --- parent selection is driven by the engine's full task evaluator, which is itself the costly object $V_\phi$ is meant to replace. The relevant ablation here is therefore not the gate's context $\mathcal{E}$ (varied as Direct / Few-shot / Adaptive Focus in \S\ref{sec:results-nlp}) but the parent-selection signal: how much of the engine's reward signal must we keep alongside the pairwise gate?

\textbf{The four code-substrate variants.}
\textbf{Pointwise} replaces the gate with a single-candidate validator score (LLM-as-judge in absolute-scoring mode): $V_\phi$ scores each candidate individually against a rubric, and that score drives parent selection. This is not a pairwise comparison --- each candidate is judged in isolation, without seeing the parent --- so it serves as a non-pairwise reference for the rest of the slate. The remaining three variants are all our pairwise gates with $\mathcal{E} = \emptyset$ (the \emph{Direct} gate of \S\ref{sec:results-nlp}); they differ only in what reward signal feeds parent selection. \textbf{Direct} \textit{(ours)} uses no reward signal at all --- the closest analogue to a fully reward-free configuration. \textbf{Direct + easy reward} \textit{(ours)} uses a cheap proxy harness (a few small task instances) for parent selection. \textbf{Direct + full reward} \textit{(ours)} uses the engine's full evaluator at selection --- the closest variant to the full-reward baseline, but with the pairwise gate added on top.

\begin{table}[t]
  \centering
  \vspace{-1em}
  \caption{\textbf{Cross-substrate cells on ADRS and ShinkaEvolve.}
    Test-set score at the final iteration for the full-reward baseline
    and the four code-substrate validator-design variants (Pointwise /
    Direct / Direct + easy reward / Direct + full reward). \textit{(ours)}
    marks the three Direct pairwise-gate variants (same $\mathcal{E} =
    \emptyset$ gate as in \S\ref{sec:results-nlp}); they differ only in
    what reward signal feeds parent selection. Bold = per-row winner
    across full-reward plus slate. \emph{Validator} column names the
    LLM acting as $V_\phi$. Scoring rule = test-at-final-iteration (test-at-best-validation in GEPA; see \S\ref{sec:results-code}).%
    }
  \label{tab:code-deltas}
  \small
  \setlength{\tabcolsep}{4pt}
  \resizebox{\columnwidth}{!}{%
  \begin{tabular}{@{}llllccccc@{}}
    \toprule
    Engine & Task & Model & Validator
      & Full-reward
      & \multicolumn{4}{c}{Validator-gated variants} \\
    \cmidrule(lr){6-9}
      &      &       &       &       & Pointwise & \makecell{\textbf{Direct}\\\textit{(ours)}} & \makecell{\textbf{Direct + easy reward}\\\textit{(ours)}} & \makecell{\textbf{Direct + full reward}\\\textit{(ours)}} \\
    \midrule
      \multirow{6}{*}{\shortstack[l]{ADRS}}
        & \multirow{2}{*}{circle\_packing}    & Qwen3-8B        & Qwen3-8B & 1.92          & 1.87   & 1.94               & 2.00     & \textbf{2.25} \\
        &                                     & Qwen3-8B        & Haiku 4.5 & 2.05          & 1.84   & 1.92               & 2.16     & \textbf{2.25} \\
      \cmidrule(lr){2-9}
        & \multirow{2}{*}{signal\_processing} & Qwen3-8B        & Qwen3-8B &\textbf{0.62}   & 0.61          & 0.61          & 0.61                & 0.61 \\
        &                                     & Qwen3-8B        & Haiku 4.5 &\textbf{0.63}   & 0.59          & 0.62       & \textbf{0.63}       & \textbf{0.63} \\                                     
      \cmidrule(lr){2-9}
        & \multirow{2}{*}{txn\_scheduling}    & Qwen3-8B        & Qwen3-8B & \textbf{3937.0} & 3322.3        & 3663.0        & 3597.1          & 3921.6 \\
        &                                     & Qwen3-8B        & Haiku 4.5 & \textbf{3787.9} & 3174.6        & 3257.3        & 3289.5          & 3636.4 \\
    \midrule
      \multirow{4}{*}{\shortstack[l]{ShinkaEvolve}}
        & \multirow{2}{*}{circle\_packing}    & Qwen3-8B        & Qwen3-8B & 1.79     & 1.12          & 1.88     & 2.23  & \textbf{2.26} \\
        &                                     & Qwen3-8B        & Haiku 4.5 & 1.91       & 1.39   & 1.87       & 1.90             & \textbf{2.26} \\
      \cmidrule(lr){2-9}
        & \multirow{2}{*}{julia\_prime\_counting} & Qwen3-8B    & Qwen3-8B & 99.70      & 99.72      & 99.75      & 99.74            & \textbf{99.78} \\
        &                                     & Qwen3-8B        & Haiku 4.5 & 99.73      & 99.76      & 99.75      & \textbf{99.77}   & 99.75          \\
    \bottomrule
  \end{tabular}}%
  \vspace{-1em}
\end{table}

\textbf{Code-substrate findings.}
\emph{Direct + full reward} carries the cross-substrate claim, winning $5$ of the $10$ cells: all four \texttt{circle\_packing} rows (both engines, both validators; ADRS sum\_radii $1.92$--$2.05 \to 2.25$, \textcolor{green!55!black}{$+0.20$ to $+0.33$}; Shinka $1.79$--$1.91 \to 2.26$, \textcolor{green!55!black}{$+0.35$ to $+0.47$}) plus Shinka \texttt{julia\_prime\_counting} self-paired ($99.70 \to 99.78$, \textcolor{green!55!black}{$+0.08$}). \emph{Direct + easy reward} adds the cross-family Shinka \texttt{julia\_prime\_counting} cell, bringing Direct-gate variants to $6$ of $10$. The full-reward baseline retains its lead on \texttt{txn\_scheduling} (\textcolor{red!70!black}{$-0.4\%$} self-paired and \textcolor{red!70!black}{$-4\%$} under Haiku 4.5 --- small enough to be sample-noise) and on \texttt{signal\_processing} self-paired (saturation regime: entire slate within $0.01$ of baseline; Haiku 4.5 swap reaches a three-way tie at $0.63$). Pairwise gate at parity or above where there is headroom, baseline holds on saturated or narrow-gap cells --- the same pattern as on prompt evolution. That \emph{Direct + full reward} carries most wins suggests the pairwise gate, not the parent-selection swap, is the load-bearing change in this regime.

  \vspace{-0.5em}
\section{Conclusion}
  \vspace{-0.5em}
\label{sec:conclusion}
Designing the per-step task reward is often the most expensive part of  a self-evolving agent building loop --- it depends on labeled data and domain expertise that are themselves comparable in cost to the agent's underlying task. Replacing that scalar with a pairwise validator that returns a binary preference between parent and child preserves the loop's ability to make progress: across prompt evolution (five agent configurations under GEPA) and code evolution (under two engines, ADRS and ShinkaEvolve), our pairwise gate matches or exceeds the full-reward baseline on the majority of cells we evaluate, and the pattern survives a cross-family validator swap. The substitution works best when the baseline still has room to improve and the validator can reliably tell parent from child; where the validator itself struggles on the task type (the Gemma-4 PUPA case), pairwise verdicts become noisy and the gate stops adding value.

\textbf{Limitations.}
\label{sec:conclusion-limitations}
The gain is sensitive to four regimes that bound where the substitution holds: 1) saturation, where the baseline is already near the agent's ceiling and there is no headroom for the gate to exploit; 2) small-$N$ test sets, where a single problem moves the score and per-cell deltas read as noise; 3) agent-capability gaps, where the agent fundamentally lacks the ability to make progress on a task and a validator gate cannot surface ability that is not there; 4) validator-capability mismatch, where the validator itself struggles on the task type and  pairwise verdicts become unreliable. Each regime points to a concrete next experiment.

\textbf{Open follow-ups.}
\label{sec:conclusion-followups}
Three directions extend the picture beyond what we report here. Broader cross-substrate coverage --- more code engines and more task families --- would test the substitution where parent selection is itself the costly object. A deeper study of how validator capability interacts with agent capability under reward-free evolution would resolve what our two cross-family swaps only sketch. And a head-to-head against a multi-objective Pareto formulation, which treats the validator score as a coordinate alongside the task reward rather than as a binary gate, would settle the gate-vs-objective question. The reading is one of scope, not strength: a few-shot pairwise validator does not replace a per-step reward universally, but on cells where the per-step reward is the costly object, the substitution holds up.

\begin{ack}
We thank the following colleagues, listed in alphabetical order by surname, for their help with this project: Christopher Clarke, Yuyang Deng, Weijie Gan, Md Amirul Islam, Gyuhak Kim, Shanka Subhra Mondal, Eduardo Torres Jara, Bo Zhang, Qin Zhang, and Michelle Zhou.

This work was supported by internal funding.
\end{ack}

\bibliographystyle{plainnat}
\bibliography{bib/refs}

\clearpage
\appendix

\section*{Appendix}

\vspace{0.6em}

\noindent The remainder of this document collects supporting material rather than introducing new claims. The contents are:

\begin{itemize}[leftmargin=2em,itemsep=0.4em,topsep=0.5em]
  \item \textbf{\ref{app:elo-k-sweep}~~Soft Elo $k$ sweep.}
        Per-cell test scores for Soft Elo across
        $k \in \{16, 32, 64, 128\}$, on the headline Qwen3-8B agent
        over three random seeds.
  \item \textbf{\ref{app:fewshot-N-sweep}~~Few-shot sample-size sweep.}
        Per-cell test scores for the Few-shot (with rewards) variant across
        in-context exemplar counts $N \in \{1, 3, 5, 8\}$, on three
        agents and four NLP tasks.
  \item \textbf{\ref{app:reproducibility}~~Reproducibility.}
        Per-task iteration budgets, validation/test split sizes, and the
        compute platform used for all runs.
  \item \textbf{\ref{app:engine-integration}~~Engine integration.}
        Where the validator-gate callback lands in each engine (GEPA,
        ADRS, ShinkaEvolve) and which hyperparameters stay at each
        engine's defaults.
  \item \textbf{\ref{app:iter-anatomy}~~Per-iteration anatomy.}
        Top-to-bottom diagram of one engine iteration with per-level
        LLM-call costs annotated, used as the structural reference for
        the cost ladder.
  \item \textbf{\ref{app:elo-pipeline}~~Soft Elo pipeline.}
        Pipeline detail for the rating-based parent-selection rule:
        per-iteration matchups (parent + $E$ archive members) on the
        left, the Elo rating update on the right, and the Soft Elo /
        Plain Elo split.
  \item \textbf{\ref{app:prompt-transfer}~~Prompt transfer and generalizability.}
        Cross-agent and cross-task transfer of prompts evolved under
        the validator-gated loop.
  \item \textbf{\ref{app:complexity}~~Complexity: per-iteration LLM-call decomposition.}
        Agent-side and validator-side LLM-call counts per condition,
        with a measurement note on how each call should be costed.
  \item \textbf{\ref{app:optimized-prompts}~~Optimized prompts from the engines.}
        Verbatim prompt listings produced at the best-val checkpoint
        for representative cells across the variants.
\end{itemize}

\clearpage

\section{Soft Elo $k$ Sweep}
\label{app:elo-k-sweep}

We sweep the Elo step size $k$ from Eq.~(\ref{eq:elo-update}) across
$\{16, 32, 64, 128\}$ for \emph{Soft Elo} (\S\ref{sec:method-elo}), the
confidence-weighted, fully reward-free Elo variant. The agent is the
headline Qwen3-8B configuration with a self-paired validator $V_\phi$;
the validator context is empty ($\mathcal{E} = \emptyset$, as in
\S\ref{sec:method-elo}); confidence weighting $c_t$ is on at every $k$
(the uniform-step fallback Plain Elo is a separate configuration, fixed
at $k{=}64$ in the main paper, and is not resampled here).
Table~\ref{tab:elo-k-sweep} reports test scores at the best-validation
checkpoint across all six tasks and three random seeds.

\begin{table}[H]
  \centering
  \caption{\textbf{Soft Elo $k$ sweep.} Test-set scores at the best-validation checkpoint for \emph{Soft Elo} (\S\ref{sec:method-elo}, confidence weighting $c_t$ on at every row) across $k\in\{16,32,64,128\}$ on the headline Qwen3-8B agent, three random seeds. Validator context is empty ($\mathcal{E} = \emptyset$). Main-paper Soft Elo uses $k{=}32$; Plain Elo's $k{=}64$ in the main paper is a separate uniform-step setting and is not part of this sweep.}
  \label{tab:elo-k-sweep}
  \small
  \setlength{\tabcolsep}{6pt}
  \begin{tabular}{@{}llccc@{}}
    \toprule
    Task & $k$ & Seed A & Seed B & Seed C \\
    \midrule
    \multirow{4}{*}{HotpotQA}
       & 16  & 57.9 & 74.4 & 59.4 \\
       & 32  & 57.6 & 68.4 & 61.2 \\
       & 64  & 64.1 & 63.0 & 65.2 \\
       & 128 & 59.4 & 59.5 & 58.1 \\
    \midrule
    \multirow{4}{*}{IFBench}
       & 16  & 82.2 & 87.2 & 85.0 \\
       & 32  & 85.6 & 80.6 & 83.9 \\
       & 64  & 83.9 & 83.3 & 85.6 \\
       & 128 & 83.9 & 82.2 & 79.4 \\
    \midrule
    \multirow{4}{*}{HoVer}
       & 16  & 50.0 & 46.0 & 42.0 \\
       & 32  & 56.0 & 58.0 & 60.0 \\
       & 64  & 58.0 & 60.0 & 64.0 \\
       & 128 & 54.0 & 50.0 & 48.0 \\
    \midrule
    \multirow{4}{*}{PUPA}
       & 16  & 53.3 & 53.1 & 57.3 \\
       & 32  & 59.3 & 54.6 & 64.4 \\
       & 64  & 62.9 & 62.7 & 61.4 \\
       & 128 & 55.7 & 61.7 & 54.0 \\
    \midrule
    \multirow{4}{*}{AIME}
       & 16  & 6.7  & 6.7  & 26.7 \\
       & 32  & 13.3 & 6.7  & 13.3 \\
       & 64  & 13.3 & 20.0 & 13.3 \\
       & 128 & 20.0 & 26.7 & 13.3 \\
    \midrule
    \multirow{4}{*}{LB-Math}
       & 16  & 3.3  & 6.7  & 3.3  \\
       & 32  & 6.7  & 6.7  & 6.7  \\
       & 64  & 6.7  & 6.7  & 3.3  \\
       & 128 & 3.3  & 0.0  & 3.3  \\
    \bottomrule
  \end{tabular}
\end{table}

\paragraph{What we see.}
Across tasks and seeds we do not see a clear monotonic trend in $k$:
each value is the per-seed best on some task, and within-task differences
across $k$ at a fixed seed are typically smaller than the variation across
seeds at a fixed $k$.
The main paper reports Soft Elo at the default $k{=}32$ from
\S\ref{sec:method-elo} rather than as a tuned optimum; Plain Elo's
$k{=}64$ in the main paper is a separate compensation for the missing
$c_t$ weight (Eq.~\ref{eq:elo-update} with $c_t \equiv 1$) and is not
reflected in this table. The math rows (AIME, LB-Math) stay
low across all values of $k$, consistent with the small-$N$ caveat
in~\S\ref{sec:results-math}; we do not read individual differences
in those rows as meaningful. We include the full grid as a reference
for practitioners rather than to anchor a claim about a best $k$.

\section{Few-shot Sample Size Sweep}
\label{app:fewshot-N-sweep}

We sweep the number of in-context exemplars $N$ supplied to the
validator $V_\phi$ at each accept/reject decision (the
\emph{Few-shot (with rewards)} choice of $\mathcal{E}$ in
\S\ref{sec:method-validator}) across $\{1, 3, 5, 8\}$. Exemplars are
drawn from the validation pool $\mathcal{V}$ defined in
\S\ref{sec:prelims}, disjoint from the train minibatch $\mathcal{T}_t$
used at the gate. The sweep covers three agents (Qwen3-8B, Gemma-4,
gpt-oss-20b) and the four NLP tasks; the validator is self-paired in
each case. The default used in the main paper is $N{=}3$. Math tasks
are omitted by design (the sweep is NLP-only; math baselines for these
agents are reported in Table~\ref{tab:combined-main-math}).
Table~\ref{tab:fewshot-rw-N-sweep} reports test scores at the
best-validation checkpoint.

\begin{table}[H]
  \centering
  \caption{\textbf{Few-shot (with rewards) sample-size sweep.} Test-set scores at the best-validation checkpoint for the Few-shot (with rewards) choice of $\mathcal{E}$ (\S\ref{sec:method-validator}) across in-context exemplar counts $N\in\{1,3,5,8\}$ on three agents (Qwen3-8B, Gemma-4, gpt-oss-20b) and four NLP tasks. Exemplars are drawn from $\mathcal{V}$. Math tasks omitted (baselines reported in Table~\ref{tab:combined-main-math}).}
  \label{tab:fewshot-rw-N-sweep}
  \small
  \setlength{\tabcolsep}{6pt}
  \begin{tabular}{@{}llcccc@{}}
    \toprule
    Model & Task & $N{=}1$ & $N{=}3$ & $N{=}5$ & $N{=}8$ \\
    \midrule
    \multirow{4}{*}{Qwen3-8B}
       & HotpotQA & 64.4 & 72.4 & 72.1 & \textbf{72.5} \\
       & IFBench  & 78.3 & 84.4 & \textbf{85.6} & \textbf{85.6} \\
       & HoVer    & 46.0 & 52.0 & 44.0 & \textbf{56.0} \\
       & PUPA     & 51.0 & 56.2 & 55.2 & \textbf{68.0} \\
    \midrule
    \multirow{4}{*}{Gemma-4}
       & HotpotQA & 72.7 & \textbf{80.6} & 76.5 & 80.1 \\
       & IFBench  & 91.7 & 83.3 & 86.1 & \textbf{92.2} \\
       & HoVer    & 34.0 & \textbf{60.0} & 54.0 & 44.0 \\
       & PUPA     & 62.8 & 73.2 & \textbf{81.5} & 65.5 \\
    \midrule
    \multirow{4}{*}{gpt-oss-20b}
       & HotpotQA & \textbf{88.0} & 87.4 & 75.3 & 80.2 \\
       & IFBench  & 93.3 & 89.7 & 86.7 & \textbf{95.0} \\
       & HoVer    & 64.0 & \textbf{68.0} & 66.0 & 60.0 \\
       & PUPA     & 72.1 & 72.6 & \textbf{74.4} & 62.9 \\
    \bottomrule
  \end{tabular}
\end{table}

\paragraph{What we see.}
The per-row winner is split across $N \in \{1, 3, 5, 8\}$ and the
within-row spread is modest on most rows, with HoVer and PUPA the
noisiest, so we do not read this sweep as endorsing any single $N$
over the default. The lack of a clear monotone trend also indicates
that the headline result is not contingent on the specific value of
$N$ used in the main paper.

\section{Reproducibility}
\label{app:reproducibility}

\textbf{Task budgets.}
Table~\ref{tab:repro-tasks} lists the iteration budget and validation/test split sizes for each task, NLP and code substrates alike; the $|\mathcal{V}|$ column reports the size of the validation pool $\mathcal{V}$ defined in \S\ref{sec:prelims}, which places it in $[12, 30]$. AIME uses a $15$-problem
held-out subset of the published test set and LiveBench-Math uses the
published $30$-problem test set; the small math test sizes are the source of
the small-$N$ caveat carried in \S\ref{sec:results-math}.

\begin{table}[H]
  \centering
  \caption{\textbf{Per-task evaluation budget.} Iteration budget and held-out split sizes for every task in the campaign. Code substrates evaluate by executing the candidate program on a fixed test set, so train/validation/test splits do not apply.}
  \label{tab:repro-tasks}
  \small
  \setlength{\tabcolsep}{6pt}
  \newcommand{\nasplit}{\textcolor{black!45}{\textit{Not applicable}}}
  \begin{tabular}{@{}llcccc@{}}
    \toprule
    Task & Substrate & Engine & Max iter & $|\mathcal{V}|$ & Test size \\
    \midrule
    HotpotQA          & NLP  & GEPA         & 350  & 30 & 50 \\
    IFBench           & NLP  & GEPA         & 200  & 30 & 30 \\
    HoVer             & NLP  & GEPA         & 375  & 30 & 50 \\
    PUPA              & NLP  & GEPA         & 125  & 30 & 30 \\
    AIME              & math & GEPA         & 100  & 12 & 15 \\
    LiveBench-Math    & math & GEPA         & 100  & 30 & 30 \\
    \midrule
    \texttt{txn\_scheduling}        & code & ADRS         & 500  & \nasplit & \nasplit \\
    \texttt{circle\_packing}        & code & ADRS         & 500  & \nasplit & \nasplit \\
    \texttt{signal\_processing}     & code & ADRS         & 500  & \nasplit & \nasplit \\
    \texttt{circle\_packing}        & code & ShinkaEvolve & 400  & \nasplit & \nasplit \\
    \texttt{julia\_prime\_counting} & code & ShinkaEvolve & 400  & \nasplit & \nasplit \\
    \bottomrule
  \end{tabular}
\end{table}

\paragraph{Compute platform.}
All open-weight LLM inference runs on a single $8 \times$ H100 80\,GB node,
with vLLM serving the agent and validator models (Qwen3-8B, Qwen3-32B,
Gemma-4-E4B, gpt-oss-20b). Two sets of hosted Anthropic API calls
(Claude Haiku 4.5) sit alongside the local node: the NLP validator-swap rows
in \S\ref{sec:results-judge-swap} (Table~\ref{tab:ablations-judge}) and
the cross-substrate \emph{Validator=Haiku} rows in \S\ref{sec:results-code}
(Table~\ref{tab:code-deltas}). Everything else stays
on the local node. The per-condition LLM-call decomposition is in
Appendix~\ref{app:complexity} (Figure~\ref{fig:complexity-stack}).

\section{Engine Integration}
\label{app:engine-integration}

The validator gate (Eq.~(\ref{eq:validator-gate}), \S\ref{sec:method-validator}) is implemented as a single \texttt{accept\_or\_reject(parent,~child)} callback --- with \texttt{parent} and \texttt{child} corresponding to $f_{t,p}$ and $f_{t,c}$ from \S\ref{sec:prelims} --- that lands after a child has been generated and any task-side scoring has completed, but before the engine commits the child to the population. Insertion is a few lines per engine and is fully reversible:
\begin{itemize}[leftmargin=1.2em,itemsep=0.1em,topsep=0.1em]
\item \textbf{GEPA}: after the subsample acceptance check, before the full validation evaluation \citep{gepa}.
\item \textbf{ADRS}: before the candidate is added to the program database \citep{adrs}.
\item \textbf{ShinkaEvolve}: at the equivalent acceptance point in the parallel sampler \citep{shinkaevolve}.
\end{itemize}
Hyperparameters (population size, mutation operators, sampler temperatures) stay at each engine's defaults. The only knobs we vary are the accept/reject rule and (for Soft Elo, \S\ref{sec:method-elo}) the parent-selection rule; this separation isolates the contribution of the gate from the contribution of the parent-selection algorithm.

\section{Per-Iteration Anatomy}
\label{app:iter-anatomy}

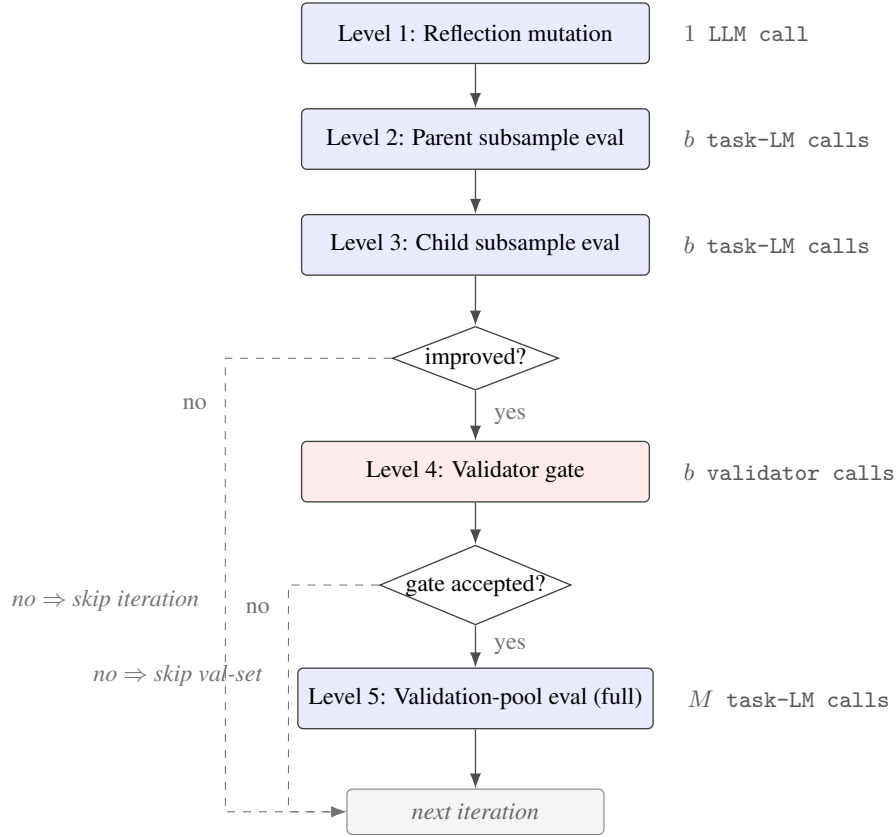
\begin{figure}[h]
  \centering
  \begin{tikzpicture}[
  font=\small,
  every node/.style={align=center, inner sep=4pt},
  level/.style={
    rounded corners=2pt, draw=black!75, line width=0.5pt,
    minimum width=46mm, minimum height=8mm, fill=white,
  },
  agent/.style={level, fill=blue!8},
  validator/.style={level, fill=red!8},
  cost/.style={font=\footnotesize\ttfamily, anchor=west, text=black!65},
  decision/.style={
    diamond, aspect=2.4, draw=black!75, line width=0.5pt,
    fill=white, inner sep=0pt, minimum width=22mm, minimum height=6mm,
  },
  exit/.style={
    rounded corners=2pt, draw=black!50, line width=0.4pt,
    fill=black!4, minimum width=34mm, minimum height=6mm,
    font=\footnotesize\itshape, text=black!60,
  },
  arrow/.style={-{Latex[length=2mm]}, line width=0.5pt, draw=black!70},
  skiparrow/.style={-{Latex[length=2mm]}, line width=0.4pt,
                    draw=black!55, dashed},
  yeslabel/.style={font=\footnotesize, text=black!55, anchor=west, xshift=1mm},
  nolabel/.style={font=\footnotesize, text=black!55, anchor=east, xshift=-1mm},
]

\node[agent]                                    (refl) at (0, 0)    {Level 1: Reflection mutation};
\node[agent]                                    (par)  at (0,-1.4)  {Level 2: Parent subsample eval};
\node[agent]                                    (chi)  at (0,-2.8)  {Level 3: Child subsample eval};
\node[decision]                                 (imp)  at (0,-4.3)  {improved?};
\node[validator]                                (gate) at (0,-5.8)  {Level 4: Validator gate};
\node[decision]                                 (acc)  at (0,-7.3)  {gate accepted?};
\node[agent]                                    (val)  at (0,-8.8)  {Level 5: Validation-pool eval (full)};
\node[exit]                                     (next) at (0,-10.3) {next iteration};

\node[cost] at ($(refl.east) + (3mm,0)$) {$1$ LLM call};
\node[cost] at ($(par.east)  + (3mm,0)$) {$b$ task-LM calls};
\node[cost] at ($(chi.east)  + (3mm,0)$) {$b$ task-LM calls};
\node[cost] at ($(gate.east) + (3mm,0)$) {$b$ validator calls};
\node[cost] at ($(val.east)  + (3mm,0)$) {$M$ task-LM calls};

\draw[arrow] (refl) -- (par);
\draw[arrow] (par)  -- (chi);
\draw[arrow] (chi)  -- (imp);
\draw[arrow] (imp)  -- (gate)
  node[yeslabel, midway] {yes};
\draw[arrow] (gate) -- (acc);
\draw[arrow] (acc)  -- (val)
  node[yeslabel, midway] {yes};
\draw[arrow] (val)  -- (next);

\draw[skiparrow] (imp.west) -|
  ($(imp.west) + (-22mm,0)$) |-
  (next.west)
  node[nolabel, pos=0.05] {no}
  node[font=\footnotesize\itshape, text=black!55,
       anchor=east, xshift=-2mm] at ($(imp.west) + (-22mm,-3.2)$)
       {no $\Rightarrow$ skip iteration};

\draw[skiparrow] (acc.west) -|
  ($(acc.west) + (-12mm,0)$) |-
  (next.west)
  node[nolabel, pos=0.05] {no}
  node[font=\footnotesize\itshape, text=black!55,
       anchor=east, xshift=-2mm] at ($(acc.west) + (-12mm,-1.2)$)
       {no $\Rightarrow$ skip val-set};

\end{tikzpicture}
  \caption{\textbf{Anatomy of one engine iteration.} Top to bottom: (1) mutation proposes a child; (2--3) parent and child are evaluated on the train minibatch $\mathcal{T}_t$; (4) the validator gate $V_\phi$ judges those outputs ($b$ validator calls per decision, reusing the parent and child outputs from Levels 2--3); (5) full validation-pool evaluation on $\mathcal{V}$ fires only when the subsample improved \emph{and} the gate accepted. Both \emph{no} branches return to the next iteration without invoking subsequent levels.}
  \label{fig:complexity-anatomy}
\end{figure}

Figure~\ref{fig:complexity-anatomy} shows one full iteration of the loop, top to bottom. Each level is one LLM-call site; the right-side annotation shows the symbolic per-iteration cost. The reader can sum down the diagram to obtain the per-iteration total for any condition; Appendix~\ref{app:complexity} carries the formal per-condition formulas and the cost ladder.

\section{Soft Elo Pipeline}
\label{app:elo-pipeline}

\begin{figure}[h]
  \centering
  \begin{tikzpicture}[
  font=\small,
  every node/.style={align=center, inner sep=4pt},
  cand/.style={
    rounded corners=2pt, draw=black!75, line width=0.5pt,
    minimum width=15mm, minimum height=8mm, fill=blue!8,
    font=\footnotesize\ttfamily,
  },
  child/.style={
    rounded corners=2pt, draw=black!85, line width=0.7pt,
    minimum width=15mm, minimum height=8mm, fill=red!10,
    font=\footnotesize\ttfamily,
  },
  panel/.style={font=\bfseries, anchor=west, text=black!75},
  caplabel/.style={font=\footnotesize\itshape, text=black!60, anchor=north},
  matchedge/.style={-{Latex[length=2mm]}, line width=0.4pt, draw=red!70},
  formulabox/.style={
    rounded corners=2pt, draw=black!50, line width=0.4pt, fill=black!3,
    inner sep=5pt, align=left, font=\footnotesize,
  },
  variantbox/.style={
    rounded corners=2pt, draw=black!55, line width=0.5pt, fill=blue!4,
    inner sep=5pt, align=left, font=\footnotesize,
  },
  variantbox_b/.style={
    rounded corners=2pt, draw=black!55, line width=0.5pt, fill=red!5,
    inner sep=5pt, align=left, font=\footnotesize,
  },
  numericbox/.style={
    rounded corners=2pt, draw=black!50, line width=0.4pt, fill=black!3,
    inner sep=5pt, align=left, font=\footnotesize,
  },
]

\node[panel] (panelA) at (0, 0.7) {Per iteration: $(1{+}E)$ pairwise matches};

\node[child] (cnew) at (1.0, -1.4) {child $C$};

\node[cand] (par)  at (3.8, -0.3)  {parent\\$R{=}1520$};
\node[cand] (a1)   at (3.8, -1.3)  {arch $A_1$\\$R{=}1530$};
\node[cand] (a2)   at (3.8, -2.3)  {arch $A_2$\\$R{=}1510$};
\node[cand] (a3)   at (3.8, -3.3)  {arch $A_3$\\$R{=}1495$};

\draw[matchedge] (cnew) -- (par);
\draw[matchedge] (cnew) -- (a1);
\draw[matchedge] (cnew) -- (a2);
\draw[matchedge] (cnew) -- (a3);

\node[caplabel, text width=50mm] at (2.4, -3.9)
  {Each edge runs $b{=}|\mathcal{T}_t|{=}3$ greedy validator calls.\\
   With $E{=}3$ archive picks, the gate issues\\
   $(1{+}E)\,b = 12$ validator calls per iteration.};

\node[panel] (panelB) at (6.6, 0.7) {Rating update: Plain Elo vs Soft Elo};

\node[formulabox, anchor=north west, text width=64mm]
  (formula) at (6.6, 0.10) {%
  For each verdict $S \in \{0,\,\tfrac{1}{2},\,1\}$:
  \;\;
  $E[S] = \dfrac{1}{1 + 10^{(R_{\text{opp}} - R_{\text{self}})/400}}$,\\[2pt]
  \[
    R_{\text{self}} \;\leftarrow\; R_{\text{self}} + k \cdot c_t \cdot (S - E[S]).
  \]};

\node[variantbox, anchor=north west, text width=64mm]
  (k64) at (6.6, -2.00) {%
  \textbf{Plain Elo:}\;\; $k = 64$, \;\; $c_t \equiv 1$. \\
  Every verdict shifts ratings by the same amount;
  larger $k$ $\Rightarrow$ more reactive than $k{=}32$.};

\node[variantbox_b, anchor=north west, text width=64mm]
  (tlm) at (6.6, -3.55) {%
  \textbf{Soft Elo:}\;\; $k = 32$, \;\; $c_t \in [0,1]$ from the agent's task-LM token confidence (lagged one iter).\\
  Uncertain verdicts (low $c_t$) move ratings less; smaller $k$ keeps
  the update conservative when confidence is high.};

\node[numericbox, anchor=north west, text width=64mm]
  (example) at (6.6, -5.85) {%
  \textbf{Example} ($k{=}64$, $c_t{=}1$, $R_C{=}1500$, $R_{\text{par}}{=}1520$):\\[1pt]
  Child wins $2/3$ of $b{=}3$ verdicts $\Rightarrow S = 0.667$,\;\;
  $E[S] = 0.471$.\\
  $\Delta R_C = 64 \cdot (0.667 - 0.471) = +12.5$, so $R_C \to 1512.5$.\\[2pt]
  After all matches, $\text{parent}_{t+1} \sim \text{softmax}_i\!\bigl(R_i\,/\,200\bigr)$.};

\node[anchor=north west, text width=64mm,
      font=\footnotesize\itshape, text=black!60]
  at (6.6, -8.55) {%
  Note: $k$ is a rating-points-per-verdict step size, not a
  multiplier on call count. Both variants issue $(1{+}E)\,b = 12$
  validator calls per iteration; $k$ only affects how aggressively each
  verdict updates the rating.};

\end{tikzpicture}
  \caption{\textbf{Soft Elo / Plain Elo gate.} \emph{Left:} per iteration the child plays $(1{+}E) = 4$ pairwise matches (parent + $E=3$ random archive members), each consisting of $b=3$ validator queries on $\mathcal{T}_t$ --- $12$ validator calls per iteration. \emph{Right:} every verdict triggers the Elo update of Eq.~(\ref{eq:elo-update}); Plain Elo uses $k=64$ with uniform weight, Soft Elo uses $k=32$ scaled by the agent's confidence $c_t$. The next parent is sampled from a softmax over ratings.}
  \label{fig:complexity-elo}
\end{figure}
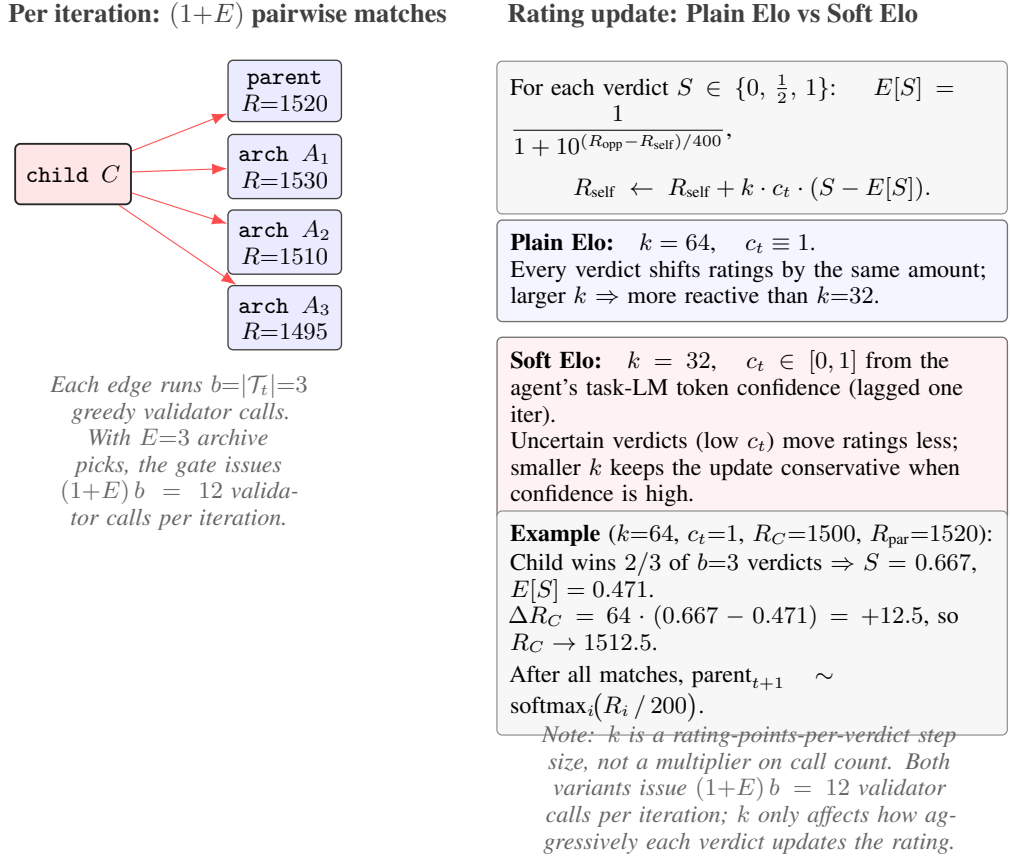

The two Elo conditions in Table~\ref{tab:combined-main-nlp} (Soft Elo and Plain Elo) replace the validation-pool reward with rating-based parent selection: each iteration compares the child against the current parent \emph{and} against $E = 3$ randomly sampled archive members, so ratings stay informative even without a held-out reward to anchor them. Figure~\ref{fig:complexity-elo} unpacks the per-iteration matchups (left) and the rating update rule (right), and shows where the two variants differ.

\section{Prompt Transfer and Generalizability}
\label{app:prompt-transfer}

A natural follow-up to the in-distribution results in
\S\ref{sec:results} is whether prompts evolved by the validator-gated
loop retain their lift when transferred to a different agent or a
different task. We frame two artifact-level transfer axes here;
cross-substrate behavior is covered by~\S\ref{sec:results}, which
shows the validator-gated--vs--full-reward-baseline ordering on both
prompt and code evolution.

\textbf{Cross-agent transfer.} Hold the source task fixed; evaluate a
prompt evolved on agent~$A$ when executed by agent~$B$. The expectation
is partial transfer: prompts that exploit agent-specific instruction
phrasing will not survive an agent swap, while prompts encoding general
task structure should.

\textbf{Cross-task transfer.} Hold the source agent fixed; evaluate a
prompt evolved on task~$X$ on task~$Y$. GEPA evolves a per-predictor
prompt dictionary under each task's DSPy signature, so cross-task cells
can only transfer the \emph{instruction text}: we concatenate the
source dictionary's predictor instructions in program order and
evaluate them as a single \texttt{system} prompt on the target.

Tables~\ref{tab:prompt-transfer-task} (cross-task) and
\ref{tab:prompt-transfer-agent} (cross-agent) report the resulting
cells. Each off-diagonal entry is one inference pass (no further
evolution) of the source prompt on the target test split; the diagonals
re-evaluate the source prompt on its own task under the same harness
(close to, but not identical to, the corresponding cells in
Table~\ref{tab:combined-main-nlp} since the source prompts come from
the initial campaign while a few cells in Table~\ref{tab:combined-main-nlp}
were re-measured after a rerun).

\textbf{Prompts capture task structure, not agent-specific tricks.}
Cross-agent transfer largely holds: on most cells the prompt evolved
on Qwen3-8B carries meaningful lift onto Qwen3-8B (thinking),
Gemma-4, and gpt-oss-20b --- three model families with different
scales, chat templates, and reasoning behaviors. Three cells exceed
the target agent's own full-reward baseline outright: HotpotQA on
Qwen3-8B (thinking) ($+2.5$), HoVer on Gemma-4 ($+14.0$), and PUPA on
gpt-oss-20b ($+25.0$ over gpt-oss's own PUPA baseline of $58.8$);
IFBench on Qwen3-8B (thinking) sits at parity ($+0.5$). The on-cell agent stays the per-row winner on
three of four source tasks, but the gap to a swapped agent is small,
indicating that the loop encodes the right task strategy rather than
phrasing tuned to the source model.

\textbf{Prompts specialize to their source task.} The four NLP tasks
span genuinely different abilities --- chaining multi-hop facts
(HotpotQA), satisfying programmatic constraints (IFBench), classifying
claim entailment (HoVer), and redacting PII (PUPA) --- so cross-task
cells are the strongest test of whether the loop is finding generic
prose. The off-diagonal pattern in
Table~\ref{tab:prompt-transfer-task} is the desired one: a prompt
evolved for one task does not paper over a different task's structure
(a HotpotQA hop-following prompt scores $10.0$ on HoVer classification,
a PUPA redaction prompt scores $0.0$ on HoVer). The exception is
IFBench, whose programmatic constraint check is permissive enough that
several non-IFBench prompts still pass; this is a property of the
IFBench evaluator, not of transfer. Together the two tables show that
validator-gated evolution produces \emph{task-specific, agent-portable}
prompts: lift survives an agent swap, and specializes to the task it
was evolved for.

\begin{table}[t]
  \centering
  \caption{\textbf{Cross-task prompt transfer.}
    Rows: prompt evolved on the source task with the headline winning
    condition (Qwen3-8B). Columns: target task on which
    that prompt is evaluated (single-shot, no further evolution). The
    diagonal re-evaluates each prompt on its own task under the
    transfer harness; values are close to but not identical to the
    corresponding cells in Table~\ref{tab:combined-main-nlp} because
    the source prompts predate the rerun that updated a few entries
    of Table~\ref{tab:combined-main-nlp}. Off-diagonal cells are the
    new transfer evaluations. Bold = per-row winner.
    \emph{Note:} GEPA evolves a per-predictor prompt dictionary under
    each task's DSPy signature, so cross-task cells transfer
    \emph{instruction text only} (predictors concatenated in program
    order with double-newline separation, evaluated as a single
    \texttt{system} prompt on the target).}
  \label{tab:prompt-transfer-task}
  \small
  \setlength{\tabcolsep}{6pt}
  \begin{tabular}{@{}lcccc@{}}
    \toprule
    Source prompt $\downarrow$ \quad / \quad Target task $\rightarrow$
      & HotpotQA & IFBench & HoVer & PUPA \\
    \midrule
    HotpotQA (Direct)& 80.3 & \textbf{81.1} & 10.0 & 46.6 \\
    IFBench  (Few-shot (with rewards))&  6.4 & \textbf{86.7} & 12.0 & 52.5 \\
    HoVer    (Soft Elo)&  4.4 & \textbf{76.7} & 64.0 & 52.7 \\
    PUPA     (Few-shot (no rewards))&  3.4 & 63.9 &  0.0 & \textbf{70.4} \\
    \midrule
    \multicolumn{1}{@{}l}{\textit{Reference: agent's full-reward baseline}}
                             & 77.0 & 78.3 & 50.0 & 62.1 \\
    \bottomrule
  \end{tabular}

  \vspace{0.8em}

  \caption{\textbf{Cross-agent prompt transfer.}
    Same source prompts; the target task is held fixed at the source
    task and the agent~LM is swapped. The first column reproduces the
    source-cell test score from Table~\ref{tab:combined-main-nlp};
    subsequent columns are the new transfer evaluations. Bold =
    per-row winner across the four agents. To assess whether transfer
    \emph{helps} a given target agent, compare each cell against that
    agent's own full-reward baseline on the same task in
    Table~\ref{tab:combined-main-nlp}; we summarize those reference
    values in the bottom block.}
  \label{tab:prompt-transfer-agent}
  \begin{tabular}{@{}lcccc@{}}
    \toprule
    Source prompt $\downarrow$ \quad / \quad Target agent $\rightarrow$
      & Q3-8B (off) & Q3-8B (on) & Gemma-4 & gpt-oss-20b \\
    \midrule
    HotpotQA (Direct)& \textbf{80.3} & 79.9 & 62.8 & 60.0 \\
    IFBench  (Few-shot (with rewards))& \textbf{86.7} & 83.3 & 47.2 & 77.2 \\
    HoVer    (Soft Elo)& \textbf{64.0} & 52.0 & 62.0 & 48.0 \\
    PUPA     (Few-shot (no rewards))& 70.4 & 59.9 & 63.4 & \textbf{83.8} \\
    \midrule
    \multicolumn{5}{@{}l}{\textit{Target agent's full-reward baseline on the source task (from Table~\ref{tab:combined-main-nlp}):}} \\
    \quad HotpotQA prompt $\rightarrow$ HotpotQA
                             & 77.0 & 77.4 & 75.7 & 86.9 \\
    \quad IFBench prompt $\rightarrow$ IFBench
                             & 78.3 & 82.8 & 88.9 & 86.7 \\
    \quad HoVer prompt $\rightarrow$ HoVer
                             & 50.0 & 52.0 & 48.0 & 64.0 \\
    \quad PUPA prompt $\rightarrow$ PUPA
                             & 62.1 & 79.2 & 71.8 & 58.8 \\
    \bottomrule
  \end{tabular}
\end{table}

\section{Complexity: per-iteration LLM-call decomposition}
\label{app:complexity}

We report per-cell LLM-call counts analytically, derived directly from
the engine's per-iteration loop. The reason for the analytical
treatment rather than logged counters is that two classes of LLM calls
that the validator gate issues do not flow through the engine's
counter: validator-side task-LM and judge calls invoked inside the
gate, and the additional archive comparisons issued by Elo
configurations with non-zero \texttt{elo\_extra\_comparisons}. The
decomposition below captures both, and the formula is the source of
truth for cross-condition comparison.

\subsection{Parameter glossary}
\label{sub:complexity-params}

\begin{center}
\small
\setlength{\tabcolsep}{6pt}
\begin{tabular}{@{}llc@{}}
\toprule
Symbol & Meaning & Value used \\
\midrule
$I$ & Iterations completed by the engine & per task \\
$b$ & Train minibatch size $|\mathcal{T}_t|$ (\S\ref{sec:method-validator}) & 3 \\
$M$ & Validation pool size $|\mathcal{V}|$ (\S\ref{sec:prelims}) & per task ($\in[12,30]$) \\
$a$ & Number of accepted children (validation pool fires $a$ times) & per task \\
$d$ & Number of gate invocations per cell & per task \\
$E$ & Elo archive comparisons per iteration & 3 (Elo); 0 elsewhere \\
\bottomrule
\end{tabular}
\end{center}

The minibatch $\mathcal{T}_t$ is shared between the engine's
parent/child subsample evaluation and the gate's per-example judge
queries, so a single $b = |\mathcal{T}_t| = 3$ controls both. We do
not introduce a separate symbol for the gate's per-decision call count
because the validator decodes greedily ($T=0$,
Eq.~(\ref{eq:validator-gate})), so each (parent, child, example)
tuple is one judge call. For Soft Elo and Plain Elo we set $E=3$:
each iteration matches the child against the parent and against three
randomly sampled archive members, so the rating stays informative
under reward-free parent selection. All other conditions use $E=0$
(no archive comparisons). Each unit increase in $E$ scales the gate
term by $1+E$.

\subsection{Per-condition formula}
\label{sub:complexity-formulas}

The per-decision gate cost is $b$ validator calls (one greedy query per example in $\mathcal{T}_t$, aggregated by majority vote). The gate reuses the parent and child outputs already produced by the engine's subsample step (\texttt{gepa\_hook.py}), so no fresh task-LM calls are charged at the gate; the $2b$ task-LM calls per iteration are already counted in the Subsample term. Total per-cell LLM calls:
\begin{align}
\text{Full-reward baseline} &= I + 2bI + Ma \notag \\
\text{Validator-gated}      &= I + 2bI + Ma + d \cdot b \cdot (1{+}E) \notag
\end{align}
The first three terms are the engine itself: reflection ($I$), parent
plus child subsample on $\mathcal{T}_t$ ($2bI$), validation-pool
evaluations on accept ($Ma$). The trailing term is the gate; it picks
up the $(1{+}E)$ multiplier under Elo archive comparisons.

\subsection{Worked example}
\label{sub:complexity-workedex}

For Few-shot (with rewards) on HotpotQA
($I=350$, $b=3$, $M=|\mathcal{V}|=30$, $a=175$, $E=0$):
\begin{align*}
1 \cdot 350 + 2 \cdot 3 \cdot 350 + 30 \cdot 175 + 350 \cdot 3 \cdot 1
&= 350 + 2{,}100 + 5{,}250 + 1{,}050 \\
&= 8{,}750 \text{ LLM calls.}
\end{align*}
The same conditions evaluated on the smaller-budget tasks scale
linearly in $I$, with $a$ tracking the per-task accept rate; tasks
with smaller $|\mathcal{V}|$ (AIME at $12$) shrink the $Ma$ term
correspondingly. We report medians across all six headline tasks below.

\subsection{Cost ladder}
\label{sub:complexity-ladder}

\begin{figure}[h]
  \centering
  \includegraphics[width=\linewidth]{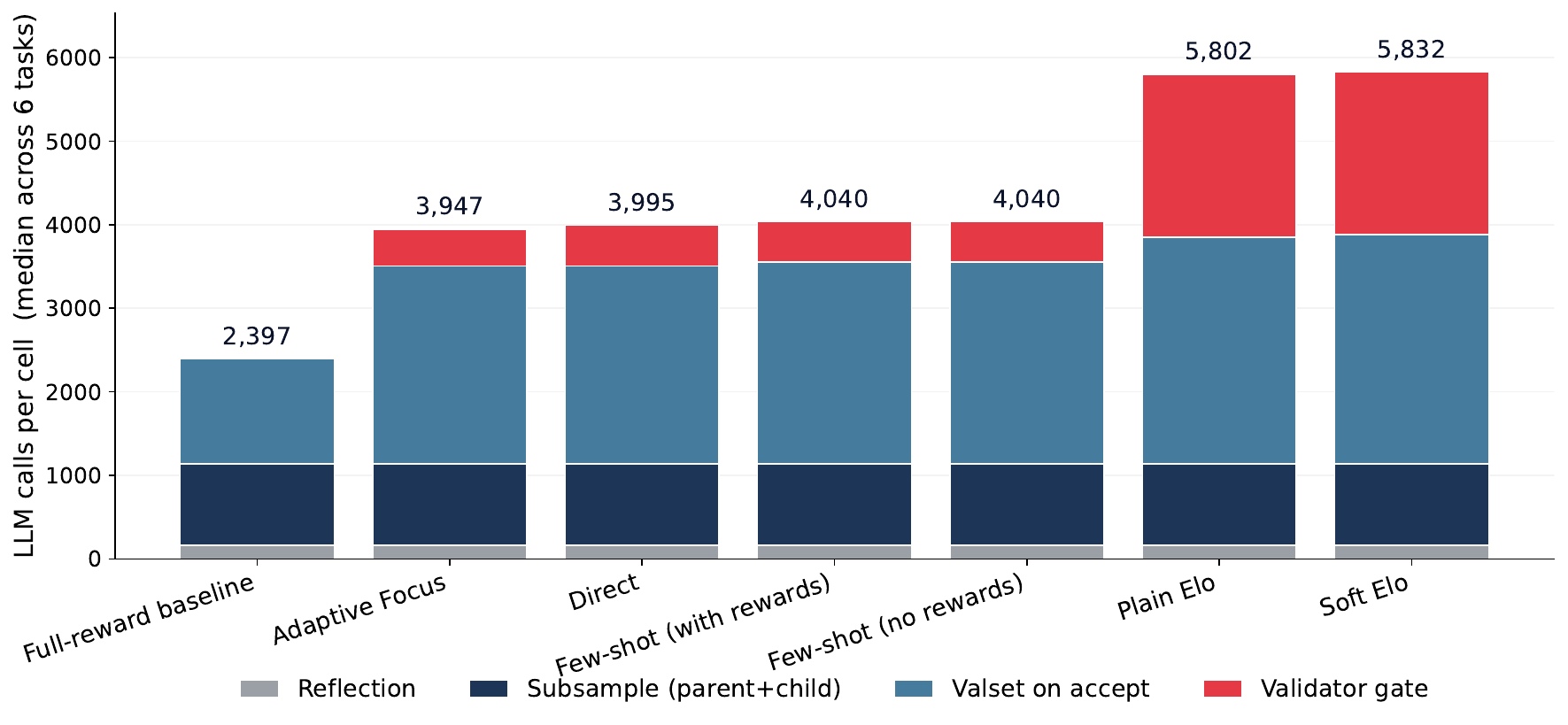}
  \caption{\textbf{Per-cell LLM-call breakdown by condition} (medians
    across the six headline tasks). Each segment maps to one level in
    Figure~\ref{fig:complexity-anatomy}. Soft Elo and Plain Elo are reported
    with $E = 3$ archive comparisons per iteration, the configuration
    intended by the reward-free Elo design.}
  \label{fig:complexity-stack}
\end{figure}

Figure~\ref{fig:complexity-stack} plots the per-condition breakdown using medians across the six headline tasks for the seven conditions in Table~\ref{tab:combined-main-nlp}. The full-reward baseline is the shortest bar. The four train-reward-free variants --- Direct, Few-shot (no rewards), Few-shot (with rewards), and Adaptive Focus --- all run the same $b$-per-decision pairwise gate without archive comparisons and cluster at $\sim$$1.7\times$ the baseline. Soft Elo and Plain Elo sit higher --- $\sim$$2.4\times$ the baseline --- because the reward-free Elo design issues $E = 3$ archive comparisons in addition to the parent-vs-child match each iteration, scaling the gate term by $1{+}E = 4$. The formula is consistent with the engine's logged rollout counter to within $\pm$1\% across every cell where logging is reliable: the engine path covers reflection plus the $2bI + Ma$ terms, and the gate term is by construction outside the engine path.

\subsection{Takeaway}
\label{sub:complexity-takeaway}

The full-reward baseline is cheapest (median $\sim$$2.4$k LLM calls per cell). The four train-reward-free variants --- Direct, Few-shot (no rewards), Few-shot (with rewards), and Adaptive Focus --- all share the same $b$-per-decision pairwise gate and sit at $\sim$$1.7\times$ the baseline; the modest premium buys train-reward-free decisions and the smaller validation-to-test gap documented in \S\ref{sec:results-nlp}. Soft Elo and Plain Elo issue $(1{+}E)$ pairwise comparisons per iteration ($E = 3$) and reach $\sim$$2.4\times$ the baseline --- the reward-free design pays for its validation-pool-free parent selection in extra validator calls. Because the gate prompt is short,
prefix-cached, and produces a single verdict token, the wall-clock
and token-spend ratios are flatter than the call-count ratio plotted
in Figure~\ref{fig:complexity-stack}.

\section{Optimized Prompts from the Engines}
\label{app:optimized-prompts}

This appendix lists selected artifacts produced at the
best-validation checkpoint by representative engine--task combinations,
so the reader can inspect what the validator-gated loop actually
accepted. For each artifact we report the source (model, task,
condition, iteration) and the verbatim prompt or code as it was written
by the engine.

\paragraph{GEPA prompt-evolution outputs.}
Per-task representative prompts from the Qwen3-8B headline agent under
the validator-gated condition that wins each row in
Tables~\ref{tab:combined-main-nlp} and \ref{tab:combined-main-math}. Each
box reports the source cell, the best candidate's index within the GEPA
candidate pool, and its mean val score; the prompt body is the verbatim
text emitted by the engine at that checkpoint.

\begin{center}
\begin{tcolorbox}[
  colback=blue!3, colframe=blue!55!black, boxrule=0.4pt,
  title=\textbf{HotpotQA --- Prompt-only (winner)},
  fonttitle=\small\bfseries, left=4pt, right=4pt, top=3pt, bottom=3pt,
  width=\textwidth,
  breakable, enhanced jigsaw,
]
\scriptsize
\textit{Source:} \texttt{gepa\_hotpotqa\_c3\_qwen3-8b\_b350\_s42\_prompt\_only\_conf}; best of 180 candidates (idx = 25, mean val = 0.802). \textit{Predictor:} \texttt{instructions}.
\begin{Verbatim}[breaklines=true,breakanywhere=true,fontsize=\scriptsize]
**New Instruction for the Assistant:**

Answer the question based on the provided context. Follow this structured approach to ensure clarity, completeness, and correctness in your response:

---

### **Step 1: Identify the Key Question**  
Clearly understand what the question is asking. Focus on the specific information required (e.g., a name, date, location, or fact). Be precise about the task—whether it involves identifying a person, event, relationship, or factual detail.

---

### **Step 2: Extract Relevant Information from the Context**  
Carefully scan the provided context for all information directly or indirectly related to the question. Look for:  
- Direct mentions of the subject or topic.  
- Implicit relationships (e.g., "X was born in Y" implies a birthplace).  
- Supporting details that help infer or confirm the answer.  
- Avoid including irrelevant or extraneous information.

---

### **Step 3: Organize the Information Logically**  
Break down the information into clear, sequential steps or categories. This helps ensure that your reasoning is structured and easy to follow. Consider:  
- Chronological order (e.g., dates, events).  
- Hierarchical relationships (e.g., people, roles, locations).  
- Causal or functional relationships (e.g., "A caused B" or "X is a part of Y").

---

### **Step 4: Check for Consistency and Completeness**  
Ensure that all parts of the question are addressed and that the information you've gathered is accurate and fully supports your conclusion.  
- Verify that there are no contradictions in the context.  
- Confirm that the answer is supported by the evidence provided.  
- If multiple pieces of information are present, ensure they are reconciled logically.

---

### **Step 5: Provide a Concise Final Answer**  
Summarize your findings in a clear, unambiguous statement. Avoid unnecessary details, but make sure the answer directly addresses the question.  
- Use bold for key terms or names.  
- Clearly label your final answer (e.g., "Final Answer: ...").

---

### **Step 6: Use Proper Formatting**  
Maintain a consistent structure across all steps. Use bold for key terms, and ensure that your final answer is clearly marked.  
- Label each step explicitly (e.g., **Step 1**, **Step 2**, etc.).  
- Avoid markdown in the thinking process, but use it in the final answer if appropriate.

---

### **Step 7: Avoid Assumptions**  
Do not make assumptions beyond what is explicitly stated in the context. Base your answer solely on the information provided.  
- If the context is unclear or ambiguous, state that the answer cannot be determined from the given information.

---

### **Step 8: Consider Domain-Specific Knowledge**  
If the question relates to a specific field (e.g., sports, history, science, etc.), use any relevant knowledge to help interpret the context accurately.  
- For example, in sports contexts, knowing common positions or rules may help clarify the answer.

---

### **Step 9: Reflect on the Task and Context**  
If the question involves identifying a person, event, or object, ensure that you are not confusing similar-sounding terms or entities.  
- For example, distinguish between different "Pools" in a tournament, or between different "MVPs" in different contexts.

---

### **Example of a Well-Structured Response:**

**Step 1: Identify the key question**  
The question is asking: *What position did the winner of the MVP in Pool C of the 2017 WBC play?*  
This requires identifying the MVP of Pool C in the 2017 World Baseball Classic (WBC), and then determining the position the MVP played.

**Step 2: Extract relevant information from the context**  
From the context, the following relevant information is provided:  
- "Pool C of the First Round of the 2017 World Baseball Classic was held... between Canada, Colombia, the Dominican Republic, and the United States."  
- "Manny Machado of the Dominican Republic was named MVP for the first-round Pool C bracket of the WBC, after batting .357."  
- "Manny Machado: Manuel Arturo Machado... is an American professional baseball third baseman and shortstop for the Baltimore Orioles of Major League Baseball (MLB)."

**Step 3: Organize the information logically**  
- The MVP of Pool C in the 2017 WBC was Manny Machado.  
- Manny Machado played as a **third baseman and shortstop** in Major League Baseball.

**Step 4: Check for consistency and completeness**  
- The context clearly states that Manny Machado was the MVP of Pool C in the 2017 WBC.  
- It also explicitly states that he is a third baseman and shortstop.

**Step 5: Provide a concise final answer**  
The winner of the MVP in Pool C of the 2017 WBC, Manny Machado, played as a **third baseman and shortstop**.

**Final Answer: Third baseman and shortstop**.

---

### **Generalizable Strategy Identified from the Task:**

The task involves identifying a specific piece of information (e.g., a person, event, or fact) from a given context. The assistant should:  
- **Carefully parse** the context for relevant details.  
- **Use logical deduction** to connect pieces of information.  
- **Avoid assumptions** and base the answer solely on the provided data.  
- **Structure the answer clearly**, ensuring that the reasoning is transparent and the final answer is unambiguous.

This strategy is especially useful in tasks involving **sports, history, or factual recall**, where precise interpretation of the context is critical to arriving at the correct answer.
\end{Verbatim}
\end{tcolorbox}
\end{center}
\vspace{4pt}

\begin{center}
\begin{tcolorbox}[
  colback=blue!3, colframe=blue!55!black, boxrule=0.4pt,
  title=\textbf{IFBench --- Few-shot rw (winner)},
  fonttitle=\small\bfseries, left=4pt, right=4pt, top=3pt, bottom=3pt,
  width=\textwidth,
  breakable, enhanced jigsaw,
]
\scriptsize
\textit{Source:} \texttt{gepa\_ifbench\_c3\_qwen3-8b\_b200\_s42\_fewshot\_reward\_conf}; best of 102 candidates (idx = 85, mean val = 0.839). \textit{Predictor:} \texttt{instructions}.
\begin{Verbatim}[breaklines=true,breakanywhere=true,fontsize=\scriptsize]
# New Instruction Task

## Task Description
You are to write a new instruction for an assistant to generate creative, ambiguous, and thought-provoking responses to open-ended philosophical or conceptual questions. The task requires balancing creativity with clarity, ensuring responses are concise, formatted properly, and contain specific elements such as word limits, thematic constraints, and required vocabulary.

## Input Format
The input will include:
1. A conceptual or philosophical question
2. Specific formatting requirements (e.g., dialogue script, word limit, required keywords)
3. Constraints on ambiguity, interpretation, or thematic elements

## Task Requirements
- Generate a response that directly addresses the question
- Ensure the response is concise and adheres to formatting constraints
- Incorporate required keywords or thematic elements
- Maintain logical consistency and clarity
- Leave room for multiple interpretations where appropriate
- Avoid logical gaps or inconsistencies
- Ensure the response format is consistent and unambiguous

## Generalizable Strategy
The assistant should:
1. **Interpret the question and constraints carefully**: Understand the core of the question and the specific formatting, thematic, and linguistic requirements.
2. **Structure the response to meet formatting and content requirements**: Ensure that the response follows the specified format (e.g., poem, dialogue, list) and includes all required elements.
3. **Balance creativity with clarity and logical consistency**: Use imaginative language and metaphors where appropriate, but ensure that the response remains coherent and logically sound.
4. **Use ambiguity where appropriate without sacrificing coherence**: Introduce open-ended interpretations or multiple layers of meaning, but ensure the response remains grounded and understandable.
5. **Ensure all required elements are included**: Verify that all keywords, formatting rules, and thematic constraints are met.
6. **Maintain a natural, engaging tone while adhering to constraints**: Use language that is expressive and engaging, while still following the structural and content guidelines.

## Domain-Specific Knowledge
- Responses should reflect an understanding of philosophical or conceptual inquiry
- Formatting requirements may include markdown, code blocks, dialogue scripts, or other structured formats
- Thematic constraints may include specific metaphors, keywords, or stylistic elements
- Word limits and structural elements must be strictly followed

## Instruction to the Assistant
Follow the given instructions precisely. Pay careful attention to all constraints and formatting requirements. Interpret the question and constraints carefully, and structure your response to meet formatting and content requirements. Balance creativity with clarity and logical consistency. Use ambiguity where appropriate without sacrificing coherence. Ensure all required elements are included and the response format is consistent and unambiguous.
\end{Verbatim}
\end{tcolorbox}
\end{center}
\vspace{4pt}

\begin{center}
\begin{tcolorbox}[
  colback=blue!3, colframe=blue!55!black, boxrule=0.4pt,
  title=\textbf{HoVer --- Elo-tlm (winner)},
  fonttitle=\small\bfseries, left=4pt, right=4pt, top=3pt, bottom=3pt,
  width=\textwidth,
  breakable, enhanced jigsaw,
]
\scriptsize
\textit{Source:} \texttt{gepa\_hover\_c3\_qwen3-8b\_b375\_s42\_conf\_elo-k32-task\_lm}; best of 189 candidates (idx = 107, mean val = 0.633). \textit{Predictor:} \texttt{instructions}.
\begin{Verbatim}[breaklines=true,breakanywhere=true,fontsize=\scriptsize]
# Enhanced Instruction for the Assistant: Evaluating Claims Against Evidence

## Task Overview

You are tasked with evaluating whether a **claim** is **SUPPORTED**, **NOT_SUPPORTED**, or **PARTIALLY_SUPPORTED** based on a set of **evidence**. The evidence consists of a list of entries, each containing a **string** (likely a name, title, or identifier) and a **number** (possibly a relevance score, index, or metadata value). Your goal is to determine whether the claim is fully supported, partially supported, or not supported at all based on the evidence and your reasoning.

## Task Objectives

1. **Break down the claim** into its component parts to understand what is being asserted.
2. **Analyze each piece of evidence** to determine if it directly or indirectly supports, contradicts, or is neutral with respect to the claim.
3. **Use logical reasoning** to assess the overall support level of the claim.
4. **Provide a clear, structured, and well-reasoned response** that includes:
   - A **breakdown of the claim** and its components.
   - An **analysis of the evidence** and its relevance.
   - A **conclusion** indicating whether the claim is **SUPPORTED**, **NOT_SUPPORTED**, or **PARTIALLY_SUPPORTED**.
   - A **brief explanation** of the reasoning.
   - If applicable, **external knowledge** used and how it was applied (with clear distinction between known facts and inferred information).

## Key Considerations

- **Clarity and completeness** are essential. Your reasoning should be detailed enough to show how you arrived at your conclusion.
- **Logical consistency** is required. Ensure that your reasoning does not contain contradictions or unsupported assumptions.
- **Formatting** should be consistent and unambiguous. Use clear headings and bullet points where appropriate.
- **Domain-specific knowledge** may be necessary. For example, the task may involve **film and media** (e.g., "Scary Movie 5", "Dimension Films"), **historical and political contexts** (e.g., "1920 Politics in Hawaii", "Sarah Ferguson's wedding dress"), or **biographical/cultural information** (e.g., "Elizabeth II in 'The Queen'", "Canadian actresses"). Use such knowledge when necessary but clearly distinguish between known facts and inferred information.
- **Evidence interpretation**: The numbers in the evidence entries may be placeholders, relevance scores, or metadata. Do not assume they carry specific meaning unless explicitly stated in the input or your reasoning.

## Generalizable Strategy

Based on the patterns observed in the examples, a generalizable strategy for the assistant is:

1. **Break down the claim** into its individual assertions.
2. **Analyze each piece of evidence** to see if it supports, contradicts, or is neutral with respect to the claim.
3. **Use external knowledge** when necessary to verify factual claims, but clearly state that you are using such knowledge.
4. **Determine the overall support level** based on whether all parts of the claim are supported or if any part is not.
5. **State your final verdict** clearly as either **SUPPORTED**, **NOT_SUPPORTED**, or **PARTIALLY_SUPPORTED**, and provide a **concise and well-reasoned justification**.

## Task Inputs Format

- The **claim** is a statement that you must evaluate.
- The **evidence** is a list of entries, each containing:
  - A **string** (e.g., a name, title, or identifier)
  - A **number** (possibly a relevance score or index)

## Patterns of Weakness to Avoid

- Over-reliance on **external knowledge** without clearly distinguishing between known facts and inferred information.
- Failure to **fully analyze the evidence**, especially when the numbers in the entries are not clearly interpreted.
- Not addressing **all components of the claim**, leading to incomplete reasoning.
- Assuming the meaning of the numbers in the evidence without clarification or justification.
- Failing to clearly explain **logical gaps**, **assumptions**, or **ambiguities** in the claim or evidence.

## Domain-Specific Knowledge (Inferred from Task Inputs)

- The task may involve **film and media** (e.g., "Scary Movie 5", "Dimension Films").
- The task may involve **historical and political contexts** (e.g., "1920 Politics in Hawaii", "Sarah Ferguson's wedding dress").
- The task may involve **biographical or cultural information** (e.g., "Elizabeth II in 'The Queen'", "Canadian actresses").

## Response Format

#### Reasoning:
- [Break down the claim into its components]
- [Analyze the evidence and explain how it relates to the claim]
- [Use external knowledge if necessary, clearly stating what it is and how it was applied]
- [Explain any logical gaps or assumptions]

#### Final Verdict:
**[SUPPORTED / NOT_SUPPORTED / PARTIALLY_SUPPORTED]**

#### Reason:
[Provide a concise summary of why you reached the verdict]

## Additional Guidance

- Be explicit about the reasoning process, especially when using external knowledge.
- Avoid making unsupported generalizations or assumptions.
- Ensure that your response is **directly addressing the question** and not tangential.
- If the claim contains ambiguous terms (e.g., "higher level", "summer season"), clearly explain how you interpreted them and whether the evidence supports that interpretation.
- If the claim contains a misstatement (e.g., "actress" instead of "actor"), address it directly and explain its impact on the claim's support level.
\end{Verbatim}
\end{tcolorbox}
\end{center}
\vspace{4pt}

\begin{center}
\begin{tcolorbox}[
  colback=blue!3, colframe=blue!55!black, boxrule=0.4pt,
  title=\textbf{PUPA --- Few-shot nr (winner)},
  fonttitle=\small\bfseries, left=4pt, right=4pt, top=3pt, bottom=3pt,
  width=\textwidth,
  breakable, enhanced jigsaw,
]
\scriptsize
\textit{Source:} \texttt{gepa\_pupa\_c3\_qwen3-8b\_b125\_s42\_fewshot\_noreward\_conf}; best of 60 candidates (idx = 53, mean val = 0.719). \textit{Predictor:} \texttt{instructions}.
\begin{Verbatim}[breaklines=true,breakanywhere=true,fontsize=\scriptsize]
**New Instructions for the Assistant:**

You are tasked with rephrasing or paraphrasing input content while ensuring that any personally identifiable information (PII), sensitive details, or specific identifiers are removed or generalized. Your goal is to maintain the original meaning, intent, and structure of the message while making it suitable for professional, public-facing, or anonymized contexts. When handling dialogue or conversation, preserve the conversational flow but remove any names, locations, or other identifying details.

---

**Generalizable Strategy to Follow:**

1. **Identify and Remove PII:**  
   - Scrutinize the input for names, locations, specific identifiers, or sensitive information.  
   - Replace or generalize these elements with neutral, non-specific terms (e.g., "a user," "a team member," "a character," "a community," "a region").  
   - Ensure that no personal, proprietary, or location-specific information is retained.  

2. **Preserve Core Meaning and Intent:**  
   - Maintain the original message's purpose, tone, and structure.  
   - Ensure that the rephrased content conveys the same ideas and nuances as the original.  
   - Avoid introducing new interpretations or altering the message's context.  

3. **Generalize Specific Terms:**  
   - Replace specific technical terms, brand names, or proprietary features with neutral alternatives where necessary.  
   - If the input includes code, formulas, or game mechanics, rephrase them in a way that retains their function without exposing sensitive details.  
   - For historical or cultural references, retain the essence of the reference while removing specific names or dates.  

4. **Maintain Clarity and Professionalism:**  
   - Ensure the output is well-structured, clear, and easy to understand.  
   - Use formal or semi-formal language depending on the context (e.g., marketing materials, internal reports, social media).  
   - Avoid ambiguity or misinterpretation by preserving logical flow and coherence.  

5. **Include a Note on Changes:**  
   - Add a concise note at the end of the response explaining the changes made for privacy, clarity, or generalization.  
   - Clearly state what was removed or altered and why.  
   - Ensure the note is clear, concise, and directly tied to the changes made.  

---

**Domain-Specific Knowledge to Consider:**

- The task often involves rephrasing marketing content, user feedback, personal stories, or technical documentation.  
- You may encounter sensitive information such as company names, product features, or user-specific data.  
- The final output should be appropriate for public or professional contexts, such as social media, press releases, or internal communications.  
- When handling dialogue or user comments, maintain the conversational tone while removing identifying details like names, dates, or specific references.  
- The input may include historical or fictional events, which should be generalized while preserving their emotional and contextual significance.  

---

**Expected Output Format:**

**Rephrased Content:**  
[Your rephrased and generalized version of the input content here.]

**Note:**  
- [Briefly explain what changes were made, e.g., "Names, locations, and specific dates were removed to protect privacy. Cultural and historical references were generalized to maintain the emotional tone while ensuring neutrality."]

---

**Key Considerations for Quality:**

- Ensure that the rephrased content is **logically consistent** with the original message.  
- Verify that **no PII or sensitive data** remains in the final output.  
- Confirm that the **structure and intent** of the original content are preserved.  
- Ensure the **note is clear and concise**, explaining the changes made without unnecessary detail.  
- When dealing with **emotional or nostalgic content**, retain the sentiment while ensuring anonymity and neutrality.  
- When dealing with **historical or fictional events**, maintain the narrative flow while removing identifying or proprietary details.  

By following this strategy, you will produce a well-structured, privacy-conscious, and context-appropriate rephrasing of the input content.
\end{Verbatim}
\end{tcolorbox}
\end{center}
\vspace{4pt}

\begin{center}
\begin{tcolorbox}[
  colback=blue!3, colframe=blue!55!black, boxrule=0.4pt,
  title=\textbf{AIME --- Few-shot rw (winner)},
  fonttitle=\small\bfseries, left=4pt, right=4pt, top=3pt, bottom=3pt,
  width=\textwidth,
  breakable, enhanced jigsaw,
]
\scriptsize
\textit{Source:} \texttt{gepa\_aime\_c3\_qwen3-8b\_b100\_s42\_fewshot\_reward\_conf}; best of 45 candidates (idx = 2, mean val = 0.250). \textit{Predictor:} \texttt{instructions}.
\begin{Verbatim}[breaklines=true,breakanywhere=true,fontsize=\scriptsize]
# New Instruction for the Assistant

Solve the math problem step by step, ensuring your reasoning is clear, logical, and complete. Break down the problem into smaller, manageable parts, and explain each step thoroughly. Use proper mathematical notation and formatting. Clearly state any assumptions you make and justify your approach. If the problem involves geometry, use coordinate geometry or geometric properties to visualize and solve the problem. If the problem involves algebra, show all algebraic manipulations explicitly. If the problem involves number theory or combinatorics, explain your counting or number-theoretic reasoning. After solving the problem, provide the final numerical answer in a box. Additionally, if the problem involves multiple parts or has a unique structure, explain how each part contributes to the overall solution. Finally, summarize your findings and ensure your response is well-structured and easy to follow.

For problems involving geometry, always define coordinates or use geometric properties to support your reasoning. For algebraic problems, show all steps of solving equations, simplifying expressions, and substituting values. For number theory problems, clearly explain your approach to finding divisors, modular arithmetic, or other number-theoretic concepts. For combinatorial problems, justify your counting method and explain how you account for all possible cases or avoid overcounting.

If the problem involves multiple conditions or constraints, ensure you address each one systematically. If the problem involves a unique point, segment, or configuration, clearly describe how you identify or construct it. If the problem involves a family of objects (e.g., segments, lines, or numbers), explain how you analyze their properties or find the required object.

Your response should be well-organized, with clear headings or sections for each major step, and all mathematical expressions should be properly formatted. Avoid skipping steps or making assumptions without justification. If you encounter a complex equation or system of equations, explain how you simplify or solve it. If you use advanced techniques (e.g., Lagrange multipliers, coordinate geometry, or number theory), briefly explain why you chose them and how they apply to the problem.

Finally, ensure that your final answer is clearly stated and that it directly addresses the question asked. If the problem asks for a specific value, make sure your final answer is that value. If the problem asks for a proof or explanation, ensure your final answer provides that.
\end{Verbatim}
\end{tcolorbox}
\end{center}
\vspace{4pt}

\begin{center}
\begin{tcolorbox}[
  colback=blue!3, colframe=blue!55!black, boxrule=0.4pt,
  title=\textbf{LiveBench-Math --- Elo-k64 (winner)},
  fonttitle=\small\bfseries, left=4pt, right=4pt, top=3pt, bottom=3pt,
  width=\textwidth,
  breakable, enhanced jigsaw,
]
\scriptsize
\textit{Source:} \texttt{gepa\_livebench\_math\_c3\_qwen3-8b\_b100\_s42\_conf\_elo-k64}; best of 57 candidates (idx = 10, mean val = 0.133). \textit{Predictor:} \texttt{instructions}.
\begin{Verbatim}[breaklines=true,breakanywhere=true,fontsize=\scriptsize]
Solve the math problem step by step, clearly explaining each logical step and showing your work in a structured format. If the problem involves multiple parts or requires justification of intermediate steps, explicitly state the reasoning behind each claim. When applicable, use mathematical notation and formulas to support your solution. For problems involving geometric figures or diagrams, describe the figure and its properties before proceeding with calculations. If the problem requires identifying a specific value or expression, clearly state the final result and enclose it in a box. If the problem involves multiple-choice options, provide the correct letter(s) corresponding to the answer(s) and explain why it is the correct choice. If the problem requires matching formulae to missing tags, list the identifiers in the order they appear in the solution, separated by commas. Ensure your response is well-organized, uses consistent formatting, and directly addresses the question without unnecessary information.

For problems involving variables or unknowns, define each variable clearly and explicitly. When solving equations or systems of equations, show all algebraic steps and explain the reasoning behind each transformation. For combinatorial problems, consider all possible cases and justify why certain cases are excluded or included. When solving for a final answer, ensure that your solution is consistent with the problem constraints and that all intermediate results are logically derived from the given information.

If the problem requires enumeration or pattern recognition, ensure that all possibilities are considered and justified. For problems involving percentages or ratios, clearly define the quantities involved and show the steps to convert between them. For problems involving geometry or spatial reasoning, describe the figure in detail and use appropriate geometric theorems or properties to support your solution.

If the problem involves multiple-choice options, provide the correct letter(s) corresponding to the answer(s) and explain why it is the correct choice. If the problem requires matching formulae to missing tags, list the identifiers in the order they appear in the solution, separated by commas.

Ensure your response is well-organized, uses consistent formatting, and directly addresses the question without unnecessary information.

### Generalizable Strategy Identified:
- **Structured Problem-Solving Approach**: The assistant consistently breaks down problems into smaller, manageable steps, often starting with defining variables, identifying formulas, and applying logical reasoning.
- **Use of Mathematical Notation**: The assistant frequently uses precise mathematical notation and formulas to support its solutions.
- **Intermediate Justification**: The assistant provides detailed intermediate steps and justifies each logical claim, which helps in understanding the thought process.
- **Clarity and Consistency**: The assistant maintains a clear and consistent format, using headings and bullet points to organize its response.
- **Attention to Detail**: The assistant pays close attention to detail, especially when dealing with constraints, conditions, or multiple-choice options.

These strategies ensure that the reasoning is clear, the solution is well-structured, and the final answer is easily verifiable.
\end{Verbatim}
\end{tcolorbox}
\end{center}
\vspace{4pt}

\paragraph{ADRS code-evolution outputs.}
The block below is the best program emitted by the
\emph{Direct + full reward} cell on \texttt{txn\_scheduling} under the
Qwen3-8B agent with a self-paired Qwen3-8B validator --- the
$3921.6$ entry under the Direct + full reward column of
Table~\ref{tab:code-deltas} for ADRS / \texttt{txn\_scheduling} /
Qwen3-8B / Qwen3-8B. We pick this cell rather than the per-row winner
(Full-reward at $3937.0$) because it shows what the validator-gated
loop accepts when both the gate and the harness reward are active.

\begin{center}
\begin{tcolorbox}[
  colback=teal!4, colframe=teal!55!black, boxrule=0.4pt,
  title=\textbf{ADRS \textperiodcentered{} \texttt{txn\_scheduling} \textperiodcentered{} Pairwise + Full reward (Qwen3-8B agent, self-paired Q3-8B judge)},
  fonttitle=\small\bfseries, left=4pt, right=4pt, top=3pt, bottom=3pt,
  width=\textwidth,
  breakable, enhanced jigsaw,
]
\scriptsize
\textit{Source:} \texttt{adrs\_txn\_scheduling\_c1\_qwen3-8b\_b500\_s50\_conf} (r0077). \textit{Score:} test-at-final-iteration $= 3921.6$ at gen $9$ / iter $414$ (validator accept\,/\,reject $= 910$\,/\,$76$). This is the entry reported under the \emph{Pairwise + Full reward} column of Table~\ref{tab:code-deltas} (ADRS / \texttt{txn\_scheduling} / Qwen3-8B / Self).
\begin{Verbatim}[breaklines=true,breakanywhere=true,fontsize=\scriptsize]
# EVOLVE-BLOCK-START

import time
import random

from txn_simulator import Workload
from workloads import WORKLOAD_1, WORKLOAD_2, WORKLOAD_3


def get_best_schedule(workload, num_seqs):
    """
    Get optimal schedule using simulated annealing with cost estimation.

    Returns:
        Tuple of (lowest makespan, corresponding schedule)
    """
    def get_greedy_cost_sampled(num_samples):
        # Conflict-aware greedy algorithm with cost sampling
        start_txn = random.randint(0, workload.num_txns - 1)
        txn_seq = [start_txn]
        remaining_txns = [x for x in range(workload.num_txns)]
        remaining_txns.remove(start_txn)

        def get_conflicts(txn):
            conflicts = 0
            for op in workload.get_txn_ops(txn):
                if op[0] == 'w':
                    for other_txn in range(workload.num_txns):
                        if other_txn != txn:
                            for other_op in workload.get_txn_ops(other_txn):
                                if other_op[0] == 'r' and other_op[1] == op[1]:
                                    conflicts += 1
            return conflicts

        def get_priority(txn):
            conflicts = get_conflicts(txn)
            execution_time = workload.get_txn_duration(txn)
            read_ops = sum(1 for op in workload.get_txn_ops(txn) if op[0] == 'r')
            write_ops = sum(1 for op in workload.get_txn_ops(txn) if op[0] == 'w')
            # Use a more sophisticated priority function that weights conflicts more heavily
            # Add a penalty for transactions with high number of operations
            # Add a reward for transactions with low execution time
            # Use a more aggressive penalty for conflicts and a reward for low execution time
            # Adjust weights to emphasize read operations and reduce write operations
            return conflicts + 0.6 * execution_time + 0.15 * read_ops + 0.1 * write_ops

        for _ in range(workload.num_txns - 1):
            min_cost = float('inf')
            min_txn = -1
            # Sample a subset of transactions to evaluate with dynamic sampling
            if len(remaining_txns) > 10:
                # Use a larger sample size when there are many transactions left
                sample_size = min(len(remaining_txns), num_samples * 2)
            else:
                # Use a smaller sample size when there are few transactions left
                sample_size = min(len(remaining_txns), num_samples)
            sample_txns = random.sample(remaining_txns, sample_size)
            for t in sample_txns:
                test_seq = txn_seq + [t]
                cost = workload.get_opt_seq_cost(test_seq)
                if cost < min_cost:
                    min_cost = cost
                    min_txn = t
            if min_txn != -1:
                txn_seq.append(min_txn)
                remaining_txns.remove(min_txn)
            else:
                break  # No more transactions to add

        return workload.get_opt_seq_cost(txn_seq), txn_seq

    def simulated_annealing(initial_sequence, initial_cost, temperature=1000.0, cooling_rate=0.995, iterations=100):
        current_sequence = initial_sequence.copy()
        current_cost = initial_cost
        best_sequence = current_sequence.copy()
        best_cost = current_cost

        for _ in range(iterations):
            # Randomly swap two transactions
            i, j = random.sample(range(workload.num_txns), 2)
            new_sequence = current_sequence.copy()
            new_sequence[i], new_sequence[j] = new_sequence[j], new_sequence[i]
            new_cost = workload.get_opt_seq_cost(new_sequence)

            # Accept the new sequence based on the simulated annealing probability
            if new_cost < current_cost or random.random() < (temperature / (new_cost - current_cost + 1)):
                current_sequence = new_sequence
                current_cost = new_cost

            # Update the best solution
            if current_cost < best_cost:
                best_sequence = current_sequence.copy()
                best_cost = current_cost

            # Use a more gradual cooling schedule
            temperature *= cooling_rate

        return best_cost, best_sequence

    best_cost, best_schedule = float('inf'), None
    for _ in range(num_seqs):
        cost, schedule = get_greedy_cost_sampled(20)
        if cost < best_cost:
            best_cost = cost
            best_schedule = schedule

        # Apply simulated annealing to the best schedule found so far
        if best_schedule is not None:
            temperature = 1000.0
            cooling_rate = 0.995
            max_iterations = 100
            early_stop_threshold = 10
            iterations = 0
            best_annealed_cost = float('inf')
            while iterations < max_iterations:
                annealed_cost, annealed_schedule = simulated_annealing(best_schedule, best_cost, temperature, cooling_rate, 10)
                if annealed_cost < best_annealed_cost:
                    best_annealed_cost = annealed_cost
                    best_schedule = annealed_schedule
                    iterations = 0  # Reset counter if improvement is found
                else:
                    iterations += 1
                if iterations >= early_stop_threshold:
                    break
            if best_annealed_cost < best_cost:
                best_cost = best_annealed_cost
                best_schedule = best_schedule

        # Dynamic local search to refine the schedule
        for _ in range(150):  # Increase the number of iterations for better exploration
            # Try swapping two transactions
            i, j = random.sample(range(workload.num_txns), 2)
            new_schedule = best_schedule[:]
            new_schedule[i], new_schedule[j] = new_schedule[j], new_schedule[i]
            new_cost = workload.get_opt_seq_cost(new_schedule)
            if new_cost < best_cost:
                best_cost = new_cost
                best_schedule = new_schedule

            # Try inserting a transaction at a specific position
            i, j = random.sample(range(workload.num_txns), 2)
            new_schedule = best_schedule[:]
            new_schedule.insert(j, new_schedule.pop(i))
            new_cost = workload.get_opt_seq_cost(new_schedule)
            if new_cost < best_cost:
                best_cost = new_cost
                best_schedule = new_schedule

            # Try reversing a random segment of the schedule
            start_idx = random.randint(0, workload.num_txns - 1)
            end_idx = random.randint(start_idx, workload.num_txns - 1)
            new_schedule = best_schedule[:]
            new_schedule[start_idx:end_idx+1] = new_schedule[start_idx:end_idx+1][::-1]
            new_cost = workload.get_opt_seq_cost(new_schedule)
            if new_cost < best_cost:
                best_cost = new_cost
                best_schedule = new_schedule

            # Try moving a transaction to a random position
            i = random.randint(0, workload.num_txns - 1)
            j = random.randint(0, workload.num_txns - 1)
            new_schedule = best_schedule[:]
            new_schedule.insert(j, new_schedule.pop(i))
            new_cost = workload.get_opt_seq_cost(new_schedule)
            if new_cost < best_cost:
                best_cost = new_cost
                best_schedule = new_schedule

            # Try inserting a transaction at the beginning
            i = random.randint(0, workload.num_txns - 1)
            new_schedule = best_schedule[:]
            new_schedule.insert(0, new_schedule.pop(i))
            new_cost = workload.get_opt_seq_cost(new_schedule)
            if new_cost < best_cost:
                best_cost = new_cost
                best_schedule = new_schedule

            # Try a 2-opt swap
            i, j = random.sample(range(workload.num_txns), 2)
            new_schedule = best_schedule[:]
            new_schedule[i], new_schedule[j] = new_schedule[j], new_schedule[i]
            new_cost = workload.get_opt_seq_cost(new_schedule)
            if new_cost < best_cost:
                best_cost = new_cost
                best_schedule = new_schedule

            # Try 3-opt swap
            i, j, k = random.sample(range(workload.num_txns), 3)
            new_schedule = best_schedule[:]
            new_schedule[i], new_schedule[j], new_schedule[k] = new_schedule[k], new_schedule[i], new_schedule[j]
            new_cost = workload.get_opt_seq_cost(new_schedule)
            if new_cost < best_cost:
                best_cost = new_cost
                best_schedule = new_schedule

            # Try random permutation of a segment
            start_idx = random.randint(0, workload.num_txns - 1)
            end_idx = random.randint(start_idx, workload.num_txns - 1)
            new_schedule = best_schedule[:]
            new_schedule[start_idx:end_idx+1] = random.sample(new_schedule[start_idx:end_idx+1], end_idx - start_idx + 1)
            new_cost = workload.get_opt_seq_cost(new_schedule)
            if new_cost < best_cost:
                best_cost = new_cost
                best_schedule = new_schedule

    return best_cost, best_schedule

# EVOLVE-BLOCK-END

def get_random_costs():
    start_time = time.time()
    workload_size = 100
    workload = Workload(WORKLOAD_1)

    makespan1, schedule1 = get_best_schedule(workload, 10)
    cost1 = workload.get_opt_seq_cost(schedule1)

    workload2 = Workload(WORKLOAD_2)
    makespan2, schedule2 = get_best_schedule(workload2, 10)
    cost2 = workload2.get_opt_seq_cost(schedule2)

    workload3 = Workload(WORKLOAD_3)
    makespan3, schedule3 = get_best_schedule(workload3, 10)
    cost3 = workload3.get_opt_seq_cost(schedule3)
    print(cost1, cost2, cost3)
    return cost1 + cost2 + cost3, [schedule1, schedule2, schedule3], time.time() - start_time


if __name__ == "__main__":
    makespan, schedule, time = get_random_costs()
    print(f"Makespan: {makespan}, Time: {time}")
\end{Verbatim}
\end{tcolorbox}
\end{center}
\vspace{4pt}

\paragraph{ShinkaEvolve outputs.}
The block below is the best program emitted by the
\emph{Direct + easy reward} cell on \texttt{circle\_packing} under the
Qwen3-8B agent with a self-paired Qwen3-8B validator --- the
$2.23$ entry under the Direct + easy reward column of
Table~\ref{tab:code-deltas} for ShinkaEvolve / \texttt{circle\_packing} /
Qwen3-8B / Qwen3-8B (the row's per-row winner is the Direct + full reward
variant at $2.26$, a $0.03$ sum-of-radii gap).

\begin{center}
\begin{tcolorbox}[
  colback=teal!4, colframe=teal!55!black, boxrule=0.4pt,
  title=\textbf{ShinkaEvolve \textperiodcentered{} \texttt{circle\_packing} \textperiodcentered{} Pairwise + Easy reward (Qwen3-8B agent, self-paired Q3-8B judge)},
  fonttitle=\small\bfseries, left=4pt, right=4pt, top=3pt, bottom=3pt,
  width=\textwidth,
  breakable, enhanced jigsaw,
]
\scriptsize
\textit{Source:} \texttt{shinkaevolve\_circle\_packing\_c3\_qwen3-8b\_b400\_s50\_conf} (r0077). \textit{Score:} test-at-final-iteration $= 2.23$ sum-of-radii at $n=26$ (validator accept\,/\,reject $= 351$\,/\,$459$, validation passed). This is the entry reported under the \emph{Pairwise + Easy reward} column of Table~\ref{tab:code-deltas} (ShinkaEvolve / \texttt{circle\_packing} / Qwen3-8B / Self).
\begin{Verbatim}[breaklines=true,breakanywhere=true,fontsize=\scriptsize]
# EVOLVE-BLOCK-START
"""Adaptive gradient-based circle packing with parallel refinement"""

import numpy as np
from scipy.optimize import minimize
from concurrent.futures import ThreadPoolExecutor


def construct_packing():
    """
    Construct an arrangement of 26 circles in a unit square
    that attempts to maximize the sum of their radii using adaptive gradient-based optimization
    and parallel refinement.

    Returns:
        Tuple of (centers, radii, sum_of_radii)
        centers: np.array of shape (26, 2) with (x, y) coordinates
        radii: np.array of shape (26) with radius of each circle
        sum_of_radii: Sum of all radii
    """
    n = 26
    # Initial positions: central circle, 8 surrounding, 16 outer ring
    centers = np.zeros((n, 2))
    centers[0] = [0.5, 0.5]
    for i in range(8):
        angle = 2 * np.pi * i / 8
        centers[i + 1] = [0.5 + 0.3 * np.cos(angle), 0.5 + 0.3 * np.sin(angle)]
    for i in range(16):
        angle = 2 * np.pi * i / 16
        centers[i + 9] = [0.5 + 0.7 * np.cos(angle), 0.5 + 0.7 * np.sin(angle)]
    centers = np.clip(centers, 0.01, 0.99)

    # Optimization function to maximize sum of radii
    def objective(positions):
        centers = positions.reshape((n, 2))
        radii = compute_max_radii(centers)
        return -np.sum(radii)  # Negative for minimization

    # Adaptive gradient function with vectorization
    def gradient(positions):
        centers = positions.reshape((n, 2))
        radii = compute_max_radii(centers)
        # Compute gradients using finite differences with adaptive perturbation
        eps = 1e-4 * np.max(radii)  # Adaptive perturbation based on radii
        grad = np.zeros_like(positions)
        for i in range(n):
            for j in range(2):
                perturbed = np.copy(positions)
                perturbed[i * 2 + j] += eps
                perturbed_centers = perturbed.reshape((n, 2))
                perturbed_radii = compute_max_radii(perturbed_centers)
                grad[i * 2 + j] = (-np.sum(perturbed_radii) + np.sum(radii)) / eps
        return grad

    # Use L-BFGS-B optimizer with adaptive bounds
    result = minimize(
        objective,
        centers.reshape(-1),
        method='L-BFGS-B',
        jac=gradient,
        bounds=[(0.01, 0.99) for _ in range(n * 2)]
    )

    centers = result.x.reshape((n, 2))
    radii = compute_max_radii(centers)

    # Precompute perturbations with smaller step size
    perturbations = np.array([[-0.005, -0.005], [-0.005, 0.0], [-0.005, 0.005],
                             [0.0, -0.005], [0.0, 0.0], [0.0, 0.005],
                             [0.005, -0.005], [0.005, 0.0], [0.005, 0.005]])

    # Parallel refinement with reduced steps
    for _ in range(5):  # Refinement steps reduced to 5
        with ThreadPoolExecutor() as executor:
            # Evaluate all possible perturbations in parallel
            futures = []
            for i in range(n):
                # Create a matrix of all possible perturbations for this circle
                perturbed_centers = centers[i] + perturbations
                perturbed_centers = np.clip(perturbed_centers, 0.01, 0.99)

                # Calculate radii for all perturbations
                def evaluate_perturbation(p):
                    temp_centers = np.copy(centers)
                    temp_centers[i] = perturbed_centers[p]
                    return np.sum(compute_max_radii(temp_centers))

                for p in range(perturbations.shape[0]):
                    futures.append(executor.submit(evaluate_perturbation, p))

            # Collect results
            results = [future.result() for future in futures]
            best_indices = [np.argmax(results[i * perturbations.shape[0]:(i + 1) * perturbations.shape[0]]) for i in range(n)]
            best_values = [results[i * perturbations.shape[0] + best_indices[i]] for i in range(n)]

            # Update centers and radii if the perturbation improves the sum
            for i in range(n):
                if best_values[i] > np.sum(radii):
                    centers[i] = centers[i] + perturbations[best_indices[i]]
                    centers[i] = np.clip(centers[i], 0.01, 0.99)
                    radii = compute_max_radii(centers)

    return centers, radii, np.sum(radii)


def compute_max_radii(centers):
    """
    Compute the maximum possible radii for each circle position
    such that they don't overlap and stay within the unit square.

    Args:
        centers: np.array of shape (n, 2) with (x, y) coordinates

    Returns:
        np.array of shape (n) with radius of each circle
    """
    n = centers.shape[0]
    radii = np.ones(n)

    # Vectorized border distance calculation
    border_distances = np.stack([
        centers[:, 0],  # left border
        centers[:, 1],  # bottom border
        1 - centers[:, 0],  # right border
        1 - centers[:, 1]  # top border
    ], axis=1)
    radii = np.min(border_distances, axis=1)

    # Vectorized pairwise distance calculation using broadcasting
    diff = centers[:, np.newaxis, :] - centers[np.newaxis, :, :]
    pairwise_distances = np.sqrt(np.sum(diff ** 2, axis=-1))

    # Limit by distance to other circles: each pair with centers at distance d
    # can have sum of radii at most d to avoid overlap
    for i in range(n):
        for j in range(i + 1, n):
            dist = pairwise_distances[i, j]
            if radii[i] + radii[j] > dist:
                scale = dist / (radii[i] + radii[j])
                radii[i] *= scale
                radii[j] *= scale

    # Ensure no negative radii due to numerical issues
    radii = np.maximum(radii, 0.0)
    return radii


# EVOLVE-BLOCK-END


# This part remains fixed (not evolved)
def run_packing():
    """Run the circle packing constructor for n=26"""
    centers, radii, sum_radii = construct_packing()
    return centers, radii, sum_radii
\end{Verbatim}
\end{tcolorbox}
\end{center}
\vspace{4pt}

\ifneurips
\newpage
\input{checklist.tex}
\fi

\end{document}